\documentclass[papersize,journal]{IEEEtran}
\usepackage{graphics} 

\usepackage[caption=false,font=footnotesize,labelfont=rm,textfont=rm,subrefformat=parens]{subfig}
\usepackage{amsmath,amsfonts}
\usepackage{amssymb}
\usepackage{algorithm}
\usepackage{mathrsfs}
\usepackage[noend]{algpseudocode}
\usepackage{multirow}
\usepackage{array}
\usepackage{booktabs}
\usepackage{threeparttable}
\usepackage{textcomp}
\usepackage{stfloats}
\usepackage{url}
\usepackage{verbatim}
\usepackage{graphicx}
\usepackage{float}
\usepackage{cases}
\usepackage{xcolor}
\usepackage{tabularx}
\usepackage{makecell}
\usepackage[table]{xcolor}

\def\BibTeX{{\rm B\kern-.05em{\sc i\kern-.025em b}\kern-.08em
    T\kern-.1667em\lower.7ex\hbox{E}\kern-.125emX}}
\usepackage{balance}

\begin{document}


\title {A Risk-Sensitive and Uncertainty-Aware Decision-Making and Control Framework for Safe and Robust Autonomous Driving}  
\author{Zhuoren Li,~\IEEEmembership{Member, IEEE}, Ran Yu, Weiqi Zhang, Ming Liu, Lu Xiong,~\IEEEmembership{Member, IEEE}, Chen Sun,~\IEEEmembership{Member, IEEE} and Bo Leng,~\IEEEmembership{Member, IEEE}
\thanks{Zhuoren Li and Ran Yu contributed equally to this work. This work has been submitted to the IEEE for possible publication. Copyright may be transferred without notice, after which this version may no longer be accessible. \textit{(corresponding author: Bo Leng).}}
\thanks{Zhuoren Li, Ran Yu, Weiqi Zhang, Ming Liu, Lu Xiong and Bo Leng are with the College of Automotive and Energy Engineering, Tongji University, Shanghai 201804, China}
\thanks{Chen Sun is with the Department of Data and Systems Engineering, The University of Hong Kong, Hong Kong}

}
\markboth{Journal of \LaTeX\ Class Files,~Vol.~X, No.~X, August~XXX}%
{Shell \MakeLowercase{\textit{et al.}}: A Sample Article Using IEEEtran.cls for IEEE Journals}

\IEEEpubid{0000--0000/00\$00.00~\copyright~2021 IEEE}

\maketitle

\begin{abstract}
Reinforcement learning (RL) has demonstrated considerable potential for autonomous driving decision-making. However, its deployment in urban autonomous driving, particularly at highly interactive unsignalized intersections, remains challenging, as learned policies may struggle to maintain both safety and robust decision-making in complex traffic situations. Conventional safety-filtering approaches typically employ fixed conservative constraints, which may improve safety at the cost of excessive intervention and degraded traffic efficiency. To address these limitations, we propose a Risk-sensitive and Uncertainty-aware Decision-making and Control (RUDC) framework for safe and robust autonomous driving. RUDC couples risk-sensitive distributional RL with ensemble-based policy uncertainty quantification, jointly accounting for tail risks in return distributions and uncertainty in learned policies. An uncertainty-aware high-order control barrier function (HOCBF)-based safety correction mechanism adaptively adjusts constraint strictness according to policy uncertainty, while a learnable residual predictor compensates for CBF model mismatches and discretization errors. Extensive simulations at unsignalized intersections demonstrate that RUDC achieves a favorable balance among safety, efficiency, and robustness, outperforming representative safe RL baselines under both nominal and challenging OOD and long-tail scenarios while satisfying real-time requirements.

\end{abstract}




\section{Introduction}
Autonomous driving (AD) has attracted sustained attention because of its potential to improve road safety, traffic efficiency, energy utilization, and driving comfort~\cite{wang2024sotifsurvey}. Despite remarkable progress in structured environments such as highways and merging ramps, achieving robust decision-making in highly interactive and safety-critical urban scenarios remains highly challenging~\cite{LiTITS2026}. Among them, unsignalized intersections are particularly challenging due to complex multi-agent interactions, and highly stochastic safety-critical events~\cite{10740674}. These characteristics increase the prevalence of out-of-distribution (OOD) and long-tail events, making it difficult to achieve both safety and efficiency~\cite{10155311, 10104197, 10534899}.

Recently, reinforcement learning (RL) has demonstrated remarkable proficiency in decision-making tasks through continuous interaction with dynamic environments~\cite{10422331, 20233614676836, 20244717409481}.However, when deployed in safety-critical driving scenarios, RL-based policies still face considerable challenges in achieving robust decision-making due to the lack of inherent safety guarantees and limited reliability assessment of learned policies~\cite{10740674}.
Safe RL has therefore become a widely adopted paradigm for safety-critical applications, aiming to maximize return while satisfying safety constraints~\cite{10675394}. A representative formulation is the constrained Markov decision process (CMDP), which restricts the expected cumulative cost below a prescribed threshold via techniques such as Lagrangian relaxation~\cite{stooke2020responsive, honari2024meta} or trust-region updates~\cite{achiam2017constrained, zhang2020first}. Nevertheless, in highly interactive driving scenarios, these approaches may suffer from sparse or weakly informative safety signals near the feasible boundary, which can hinder learning efficiency and policy stability.
\IEEEpubidadjcol

An alternative line of research improves safety by correcting the actions proposed by an RL policy through action masking~\cite{shixin2024unmanned}, safe projection~\cite{dalal2018safe, zhang2023evaluating}, or safety energy functions~\cite{10402567,10610959, 9718195}. While action masking effectively filters out hazardous maneuvers by restricting network updates to a predefined safe action set, its applicability is largely confined to discrete action spaces. Safe projection techniques, on the other hand, actively map unsafe exploratory actions back into a safe set via linearization or gradient descent. Nonetheless, their success relies heavily on accurate risk estimations. Among safety energy functions, control barrier functions (CBFs) or high-order CBFs (HOCBFs) are widely used to enforce safety constraints at execution time \cite{8796030,xiao2022hocbf,xiong2023dtcbf}. However, since the reliability of RL decisions is not explicitly considered, existing safety filters usually employ fixed conservative constraints to guarantee safety, which may limit the robustness and adaptability of autonomous driving policies in highly interactive scenarios. In highly interactive scenarios such as unsignalized intersections, this can lead to premature braking, excessive yielding, and reduced traffic efficiency. More fundamentally, the same intervention is applied to both confident and uncertain decisions, even though their safety implications can differ substantially.

In OOD and long-tail safety-critical driving scenarios, the challenge lies not only in addressing unreliable policy decisions but also in avoiding excessive safety conservatism. Without explicit uncertainty estimation, an RL agent may become overconfident in unfamiliar situations, resulting in unsafe behaviors, whereas fixed safety filters may over-constrain reliable decisions and compromise operational efficiency~\cite{10104197}.
Quantifying such uncertainty therefore provides a principled basis for reliability-aware control, enabling stronger intervention when the policy is unreliable while preserving flexibility when it is trustworthy~\cite{10155311, 10073955}. In general, RL decision uncertainty can be decomposed into epistemic uncertainty (EU), which stems from limited data coverage, and aleatoric uncertainty (AU), which arises from inherent environmental stochasticity~\cite{hullermeier2021uncertainty}. Recent autonomous driving studies have therefore begun to incorporate uncertainty into RL-based decision-making~\cite{10155311, 10534899, 10073955, 10107652, caozhongTISconfidence}. However, many existing methods use uncertainty only as a threshold-triggered signal for activating or switching to a predefined backup policy~\cite{10107652, caozhongTISconfidence}. Such mechanisms are often scenario-dependent and sensitive to manually chosen thresholds, limiting their adaptability and generalizability.

Beyond uncertainty awareness, robust safe decision-making also requires explicit consideration of tail risks, since rare but catastrophic outcomes may be overlooked by expectation-based RL objectives. In these settings, the consequences of decision errors are inherently asymmetric, as infrequent adverse events can lead to disproportionately large safety and efficiency losses. Since standard RL typically optimizes the expectation over the return distribution, it may favor behaviors that perform well on average yet remain vulnerable under adverse conditions, thereby making safety assurance difficult.

To address these limitations, we propose a unified \textbf{R}isk-sensitive and \textbf{U}ncertainty-aware \textbf{D}ecision-making and \textbf{C}ontrol (RUDC) framework. This framework intrinsically couples risk-sensitive distributional reinforcement learning with policy uncertainty quantification. By adaptively adjusting the strictness of formal safety constraints according to the estimated policy uncertainty, RUDC achieves a better balance between safety and efficiency during navigation through unsignalized intersections, enhancing robustness under challenging OOD and long-tail scenarios. This article is an extension of our preliminary work~\cite{11423819}, which constructed a basic risk-sensitive distributional critic architecture to generate risk-averse policies and employed a HOCBF as a safety filter to rectify the nominal RL actions. The specific extensions and contributions are summarized as follows:
\begin{itemize}
\item \textbf{Unified Quantification of Tail-Risk and Uncertainty:} We present a novel unified decision-making formulation that intrinsically couples risk-sensitive distributional RL with ensemble-based policy uncertainty quantification. By jointly considering tail risks in return distributions and uncertainty in learned policies, the proposed framework improves decision reliability and robustness in safety-critical driving scenarios.

\item \textbf{Uncertainty-Aware Adaptive Safety Correction:} We develop an uncertainty-aware HOCBF-based safety correction mechanism to refine nominal RL actions. By adaptively adjusting constraint strictness based on policy uncertainty and compensating CBF model mismatches and discretization errors through a residual predictor, it mitigating excessive conservatism while maintaining safety guarantees, and enabling a better balance between exploratory flexibility and safe operation.

\item \textbf{Extensive Benchmarking Validation and OOD Case Analysis:} Extensive comparisons with leading Safe RL baselines and ablation studies demonstrate the effectiveness of RUDC in balancing safety and efficiency in highly interactive traffic scenarios. Further case analyses under challenging OOD and long-tail scenarios reveal how the framework adaptively adjusts safety interventions according to policy uncertainty.

\end{itemize}

\section{Preliminaries}
\subsection{Distributional Reinforcement Learning}
\label{sec:distributinoal_rl}
The decision problem can be modeled as an Markov Decision Process (MDP) $\mathcal{M} = (\mathcal{S}, \mathcal{A}, \mathcal{P}, r, \rho_0, \gamma)$,
where $\mathcal{S}$ and $\mathcal{A}$ are the state and action spaces, $\mathcal{P}:\mathcal{S}\times\mathcal{A}\times\mathcal{S}\to[0,1]$ is the transition probability function, $r:\mathcal{S}\times\mathcal{A}\to\mathbb{R}$ is the reward function, $\rho_0$ is the initial-state distribution, and $\gamma\in(0,1)$ is the discount factor. A stochastic policy $\pi:\mathcal{S}\to{\mathcal{P}}(\mathcal{A})$ maps each state to a probability distribution over actions. The set of all policies is defined as $\Pi$. The goal of standard RL is to maximize the cumulative discounted reward, given by:
$\mathcal{L}(\pi)
=
\mathbb{E}_{\pi,\mathcal{P}}\!\left[
\sum_{t=0}^{\infty}
\gamma^t
r(s_t,a_t)
\right]$. Given a policy $\pi \in \Pi$, its \textit{action-value function} $Q^\pi$ is defined as the expected discounted return: $Q^\pi(s,a) = \mathbb{E} [\sum_{t=0}^{\infty} \gamma^t r_t | s_0=s, a_0=a]$. The corresponding \textit{Bellman optimality operator} $\mathcal{T}^{*}$ is defined as:
\begin{equation}
\begin{aligned}
\mathcal{T}^{*}Q(s,a) &:= \mathbb{E}[r(s,a)] + \gamma \mathbb{E}_{s^{\prime} \sim \mathcal{P}} \left[ \max_{a^{\prime} \in \mathcal{A}} Q(s^{\prime},a^{\prime}) \right]
\end{aligned}
\label{eq:standard_bellman}.
\end{equation}

Unlike standard RL, distributional RL models the return distribution $Z(s,a)$, whose expectation corresponds to the action-value function $Q(s,a) = \mathbb{E}[Z(s,a)]$. Then the \textit{distributional Bellman optimality operator} $\mathcal{T}_{D}^{*}$ are defined as:
\begin{equation}
\mathcal{T}_{D}^{*}Z(s,a) \overset{D}{:=} r(s,a) + \gamma Z(s^{\prime},a^{*})
\label{eq:dist_bellman_operators},
\end{equation} 
where $s^{\prime} \sim \mathcal{P}, \ a^{*} = \arg\max_{a^{\prime} \in \mathcal{A}} \mathbb{E}\left[ Z(s^{\prime},a^{\prime}) \right]$, $\overset{D}{:=}$ indicates that the random variables on both sides of the equation share the same probability distribution.

To parameterize and approximate the return distribution $Z(s,a)$, Quantile Regression (QR) \cite{dabney2018distributional} is commonly employed. Let $F_Z(z) = \mathbb{P}(Z \le z)$ denote the cumulative distribution function (CDF) of the random variable $Z$. The quantile function $F_Z^{-1}$ can be expressed as the inverse of the CDF. Given quantile fraction $\tau$, we have $F_Z^{-1}(\tau) := \inf\{z \in \mathbb{R} : \tau \le F_Z(z)\}$. Following \cite{ma2025dsac}, we approximate $F_Z^{-1}$ by defining a set of discrete quantile fractions $\{\tau_i\}_{i=0}^{N_q}$ within the interval $[0,1]$. By utilizing the midpoints $\hat{\tau}_i = (\tau_i + \tau_{i+1})/2$ of adjacent fractions, the quantile function can be efficiently represented and learned.

\subsection{Control Barrier Functions}
Consider an input-affine control system:
\begin{equation}
\begin{aligned}[b]
\dot {\boldsymbol{x}} = f(\boldsymbol{x}) + g(\boldsymbol{x})\boldsymbol{u},
\label{eq:con_affine_sys}
\end{aligned}
\end{equation} where $\boldsymbol{x}\in\mathcal{X}\subset\mathbb{R}^n$ is the system state, $\boldsymbol{u}\in\mathcal{U}\subset\mathbb{R}^{u_n}$ is the control input, and $f$ and $g$ are locally Lipschitz. In safety-critical scenarios, the goal is to keep the system state within a safe region, formalized as a forward invariant set.
\newtheorem{definition}{Definition}
\begin{definition} [Forward invariant set]
The set $\mathcal{C}$ is forward invariant for system (\ref{eq:con_affine_sys}) if for every initial condition $\boldsymbol{x}(t_0)\in \mathcal{C}$, $\boldsymbol{x}(t)\in \mathcal{C}$ for $\forall t\geq t_0$. For a continuously differentiable function $h:\mathcal{X}\rightarrow \mathbb{R}$, let
\begin{equation}
\begin{aligned}[b]
\mathcal{C}:=\{\boldsymbol{x}\in\mathcal{X}:h(\boldsymbol{x})\geq0\}, \\ Int(\mathcal{C}):=\{\boldsymbol{x}\in\mathcal{X}:h(\boldsymbol{x})>0\}, \\\partial\mathcal{C}:=\{\boldsymbol{x}\in\mathcal{X}:h(\boldsymbol{x})=0\}.
\label{eq:safe_set_define}
\end{aligned}
\end{equation}
\end{definition}

\begin{definition} [CBF \cite{8796030}] 
Let a class $\mathcal{K}$ function be a function $\alpha:[0,a)\to[0,\infty),a>0$ that is strictly increasing and with $\alpha(0)=0$. Given the superlevel set $\mathcal{C}$ as in (\ref{eq:safe_set_define}), $h$ is a CBF if there exists a class $\mathcal{K}$ function $\alpha$ such that
\begin{equation}
\begin{aligned}[b]
\sup_{\boldsymbol{u}\in\mathcal{U}}\left[L_{f}h(\boldsymbol{x})+L_{g}h(\boldsymbol{x})\boldsymbol{u}\right]\geq-\alpha(h(\boldsymbol{x})), \forall \boldsymbol{x}\in\mathcal{C},
\label{eq:cons_cbf}
\end{aligned}
\end{equation}
\end{definition} where $L_{f}h,L_{g}h$ denote the Lie derivatives along $f$ and $g$, respectively. Since decision and control algorithms for autonomous vehicles (AVs) are typically calculated and executed in discrete time, we consider the corresponding discrete-time system $\boldsymbol{x}_{k+1}=f(\boldsymbol{x}_k)+g(\boldsymbol{x}_k)\boldsymbol{u}_k$, where $k\in\mathbb{N}$ denotes the time step. A continuous function $h(x)$ is a valid discrete-time CBF if there exists a class $\mathcal{K}$ function $\alpha$ (with $\alpha(x) \le x$) such that:
\begin{equation}
\begin{aligned}[b]
\sup_{\boldsymbol{u}\in\mathcal{U}}\left[\Delta h(\boldsymbol{x},\boldsymbol{u})+\alpha(h(\boldsymbol{x}))\right]\geq0, \forall \boldsymbol{x}\in\mathcal{C},
\label{eq:dis_cbf}
\end{aligned}
\end{equation} where $\Delta h(\boldsymbol{x}_k,\boldsymbol{u}_k)=h(\boldsymbol{x}_{k+1})-h(\boldsymbol{x}_{k})$. Note that linear class $\mathcal{K}$ function is commonly used in the discrete domain, i.e., $\alpha(h(\boldsymbol{x}))=\lambda h(\boldsymbol{x}), \lambda\in(0,1]$. Consequently, it follows that $h(\boldsymbol{x}_k) \ge (1 - \gamma)^k h(\boldsymbol{x}_0)$ for all $k \in \mathbb{N}$. Provided that $\gamma \in (0, 1]$, the forward invariance of the safe set is rigorously preserved. Under this formulation, the standard discrete-time CBF naturally reduces to the discrete-time exponential CBF~\cite{xiong2023dtcbf}.

\section{Methodologies}
\begin{figure}
    \centering
    \subfloat{%
        \includegraphics[width=0.48\textwidth]{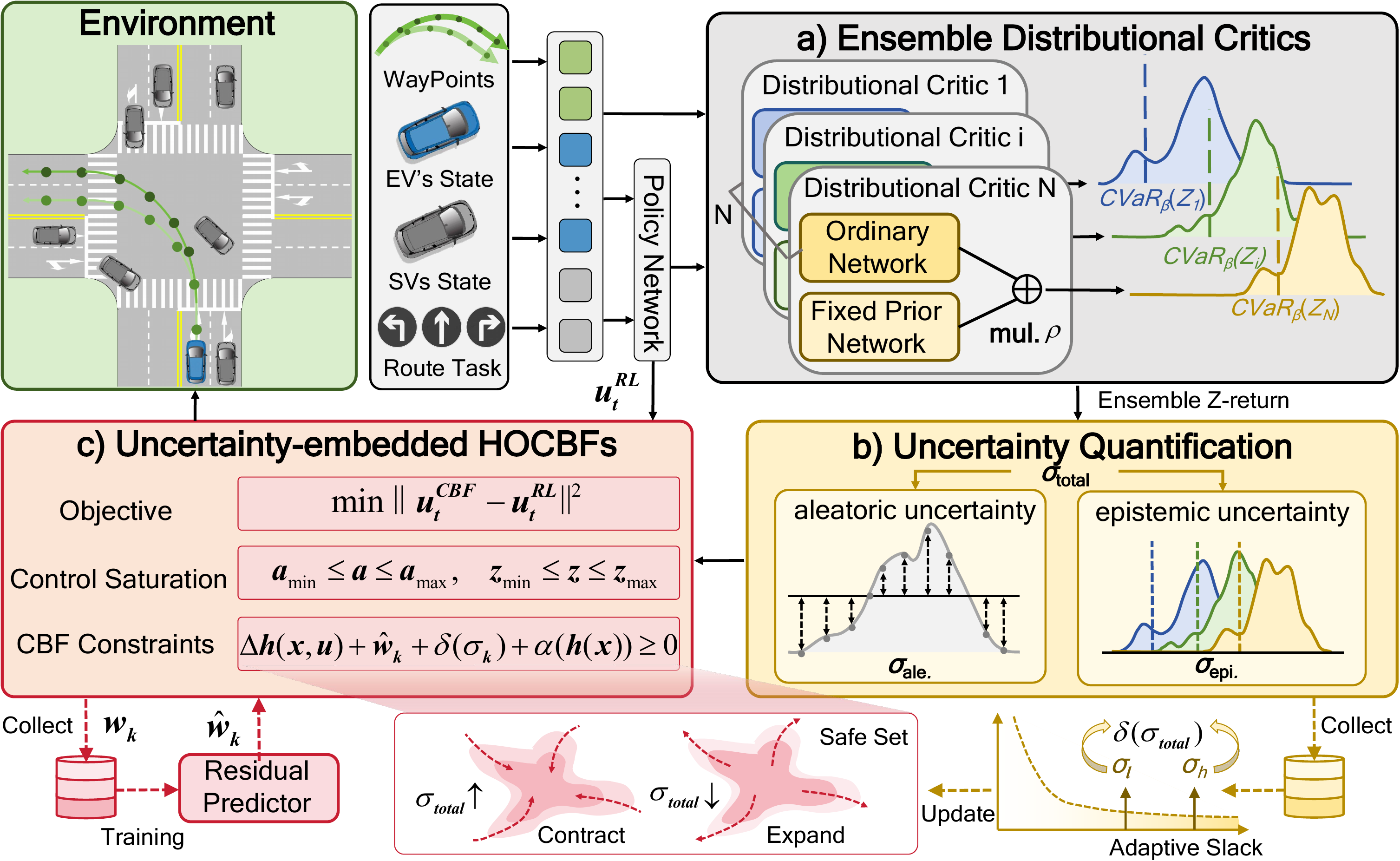}
        \label{framework}
    }
    \caption{Framework architecture. a) Ensemble critics predict return distributions to facilitate a risk-averse evaluation of the driving policy. b) Total uncertainty is quantified by decoupling epistemic (ensemble divergence) and aleatoric (distribution variance) components to dynamically modulate the adaptive slack term. c) Uncertainty-embedded HOCBFs rectify nominal actions, balancing safety-critical requirements with operational flexibility based on quantified reliability.}
    \label{fig:framework}
\end{figure}
\subsection{Uncertainty-aware Risk-sensitive Reinforcement Learning}
\subsubsection{Risk-sensitive Reinforcement Learning}
In interactive driving scenarios, stochastic outcomes arise from the unpredictable behaviors of surrounding agents. Standard RL optimizes the expected return $\mathbb{E}[Z(s,a)]$, which focuses on average performance but often ignores the tail risks of safety-critical tasks. In these scenarios, rare but catastrophic events create highly asymmetric or multimodal return distributions. Consequently, a risk-neutral policy with a high expected return may still assign non-negligible probability to unsafe outcomes, necessitating a risk-sensitive objective to suppress low-return tail risks.

To explicitly account for this tail risk, we adopt a risk-sensitive distributional RL formulation. A risk-sensitive policy is obtained by maximizing the distorted expectation of $Z(s,a)$:
\begin{equation}
\begin{aligned}
a^*(s_t)
&= \arg\max_{a \in \mathcal{A}} \Xi\!\left(\hat\rho_{DE}[Z(s,a)]\right) 
\\&\approx \arg\max_{a \in \mathcal{A}} \sum_{i=0}^{N_q-1} (\tau_{i+1}-\tau_i)\,\zeta'(\hat{\tau}_i)\, Z_{\hat{\tau}_i}(s_t,a),
\end{aligned}
\label{eq:risk_sensitive_policy}
\end{equation}
where $\zeta(\cdot):[0,1]\rightarrow[0,1]$ denotes the distortion function, which is strictly increasing and satisfies $\zeta(0)=0$ and $\zeta(1)=1$. $\Xi(\hat\rho_{DE}[Z])$ 
is the distorted expectation of distribution $Z$ under the distortion function $\zeta(\cdot)$. When $\zeta(\tau)=\tau$, $\Xi(\hat\rho_{DE}[Z])$ reduces to the standard expectation and the resulting policy is risk-neutral. Conversely, a risk-averse policy is achieved when $\zeta(\tau)$ assigns greater weight to the lower tail of the return distribution.

In this work, we employ the Conditional Value-at-Risk (CVaR) \cite{rockafellar2002cvar} to formalize this risk aversion. CVaR quantifies the expected performance under worst-case outcomes. Given a confidence level $\beta\in(0,1]$, let $\operatorname{VaR}_{\beta}(Z)=F_Z^{-1}(\beta)$ denote the $\beta$-quantile of $Z$. The CVaR is defined as:
\begin{equation}
\begin{aligned}
\operatorname{CVaR}_{\beta}(Z)=\mathbb{E}\left[ Z \mid Z \le \operatorname{VaR}_{\beta}(Z)\right]=\frac{1}{\beta}\int_{0}^{\beta}F_Z^{-1}(\tau)\,d\tau.
\end{aligned}
\label{eq:cvar}
\end{equation} 

This measure corresponds to applying a specific distortion function, $\zeta_{\beta}(\tau)=\min\left\{{\tau}/{\beta},\,1\right\}$. Consequently, the final risk-sensitive policy is optimized via:
\begin{equation}
\begin{aligned}
a^{*}(s)=\arg\max_{a\in\mathcal{A}} \operatorname{CVaR}_{\beta}(Z(s,a)).
\end{aligned}
\label{eq:cvar_policy}
\end{equation}

\subsubsection{Ensemble Learning and Uncertainty Quantification}
To quantify the reliability of the learned policy, we employ deep ensembles~\cite{ganaie2022ensemble} for uncertainty estimation. For robust OOD recognition, the ensemble is expected to produce consistent performance on in-distribution samples while preserving sufficient diversity on OOD data~\cite{rame2021dice}. To this end, we construct an ensemble critic architecture composed of $N_{\bf{ens.}}$ critics $Z_n^{\theta}$ and their corresponding target critics $Z_n^{\bar{\theta}}$, where $n \in \{1,\ldots,N_{\bf{ens.}}\}$. To structurally enforce diversity, Bootstrapping~\cite{osband2016bootstrapped} is utilized, ensuring each ensemble member accesses a unique subset of the experience replay buffer. Furthermore, we integrate a Randomized Prior Function (RPF)~\cite{osband2018prior} by adding a fixed prior network (FPN) of the same architecture to each ensemble member. This enhances Bayesian posterior estimates and prevents the critics from collapsing to identical predictions in unfamiliar state spaces. The $Z$-return of the $n$-th ensemble member is formulated as:
\begin{equation}
\begin{aligned}[b]
Z_{n,\tau}^{\theta}(s,a)=\frac{\mathcal{O}(s,a;\theta_n) +\rho\mathcal{F}(s,a;\tilde{\theta}_n)}{1+\rho},
\label{eq:fpn}
\end{aligned}
\end{equation} where $\mathcal{O}(\cdot)$ and $\mathcal{F}(\cdot)$ denote the original trainable network and the FPN, respectively. The parameters $\tilde{\theta}_n$ of the FPN are frozen, and $\rho$ represents the prior scaling factor. Then, the temporal difference (TD) error of $n$-th critic is defined as: 
\begin{equation}
\begin{aligned}[b]
\delta _{ij}^n = r + \gamma \left[ {\bar Z_{\widehat {{\tau _i}}}^{\bar \theta }(s',a') - \alpha_{\bf{tem}} \log {\pi _{\bar \phi }}(a'|s')} \right] - Z_{n,\widehat {{\tau _j}}}^\theta (s,a),\label{eq:td_ensemble}
\end{aligned}
\end{equation} where $\bar{\theta}, \theta,$ and $\bar{\phi}$ represent the parameters of the target critic network, critic network, and target actor network, respectively. $\alpha_{\bf{tem}}$ is the temperature parameter. $\bar Z_{\widehat {{\tau _i}}}^{\bar \theta }(s',a') = \frac{1}{N_{\bf{ens.}}}\sum_{n = 1}^{N_{\bf{ens.}}} {\bar Z_{n,\widehat {{\tau _i}}}^{\bar \theta }(s',a')}$ is the average $Z$-return of the $N_{\bf{ens.}}$ target critics, with $a'\sim\pi_{\bar{\phi}}(s')$. By incorporating the quantile Huber loss~\cite{huber1992robust} and a Bernoulli bootstrap mask $m_{n}\sim Bernoulli(\mathrm{p})$ for $\mathrm{p}\in(0,1]$, the objective function for training the ensemble critics is comprehensively formulated as:
\begin{equation}
\begin{aligned}[b]
\mathcal{L}^{n}_Z(\theta)=\frac{1}{|\mathcal{B}|}\sum_{(s,a,r,s')\in\mathcal{B}}\sum_{i=0}^{N_q-1}\sum_{j=0}^{N_q-1}m_{n}(\tau_{i+1}-\tau_i)\rho_{\hat{\tau}_j}^\kappa\left(\delta_{ij}^{n}\right),
\label{eq:ensemble_huber_loss}
\end{aligned}
\end{equation} where $\mathcal{B}$ represents a mini-batch of transitions sampled from the replay buffer. The function $\rho_{\tau}^{\kappa}(\delta_{ij})$ denotes the quantile Huber loss evaluated at a specific quantile fraction $\tau$ with a threshold parameter $\kappa$, formulated as:
\begin{equation}
\begin{aligned}
\rho_{\tau}^{\kappa}(\delta_{ij}) &= \left|\tau - \mathbb{I}\left\{\delta_{ij} < 0\right\}\right| \frac{\mathcal{L}_{\kappa}(\delta_{ij})}{\kappa}, \\
\mathcal{L}_{\kappa}(\delta_{ij}) &= 
\begin{cases}
\frac{1}{2}\delta_{ij}^{2}, & \text{if } |\delta_{ij}| \leq \kappa \\
\kappa\left(|\delta_{ij}| - \frac{1}{2}\kappa\right), & \text{otherwise}
\end{cases}
\end{aligned}.
\label{eq:quantile_huber_combined}
\end{equation}

The actor objective is correspondingly modified to maximize the risk-averse return across the ensemble:
\begin{equation}
\begin{aligned}[b]
\mathcal{L}_\pi(\phi)=\mathbb{E}_{s\sim \mathcal{D},a\sim\pi_\phi}[\alpha_{\bf{tem}}\log(\pi_\phi(a|s))-\overline{\operatorname{CVaR}}_{\beta}(s,a)],
\label{eq:actor_loss_ensemble}
\end{aligned}
\end{equation} where $\overline{\operatorname{CVaR}}_{\beta}(s,a)=\frac{1}{N_{\bf{ens.}}}\sum_{n=1}^{N_{\bf{ens.}}}CVaR_{\beta}(Z^{\theta}_{n})$ and $\mathcal{D}$ denotes the replay buffer.

Based on the Law of Total Variance~\cite{sale2023second}, we decompose the total uncertainty arising in the decision-making process into AU $\mathscr{U}_{\text{ale.}}$ and EU $\mathscr{U}_{\text{epi.}}$. Let $Z$ denote the stochastic return of a state-action pair $(s,a)$, and $M$ represent the latent dynamic model or environmental parameters. Accordingly, the overall variance can be decomposed as:
\begin{equation}
\begin{aligned}[b]
\operatorname{Var}(Z) = \mathbb{E}_M[\operatorname{Var}(Z|M)] + \operatorname{Var}_M(\mathbb{E}[Z|M]).
\end{aligned}
\end{equation}

Then,  the total uncertainty $\mathscr{U}_{\text{total}}$ can be expressed as:
$\mathscr{U}_{\text{total}} = \mathscr{U}_{\text{ale.}} + \mathscr{U}_{\text{epi.}}$. In decision-making scenarios focusing on extreme risks, we extend the above variance decomposition logic to the lower $\beta$-tail of the return distribution. EU quantifies the degree of disagreement among different ensemble members regarding their estimates of tail expectations, and the standard deviation of EU $\sigma_{\text{epi.}}$ is defined as:
\begin{equation}
\begin{aligned}[b]
\sigma_{\text{epi.}} = \sqrt{ \operatorname{Var}_{n=1}^N \Big( \operatorname{CVaR}_\beta(Z_n^{\theta}(s,a)) \Big) }.
\end{aligned}
\end{equation}

AU measures the average level of intrinsic fluctuation within each model's own tail distribution, i.e., the expected conditional variance when the distribution falls below the VaR. Its standard deviation $\sigma_{\text{ale.}}$ is defined as:
\begin{equation}
\begin{aligned}[b]
\sigma_{\text{ale.}} = \sqrt{ \frac{1}{N_{\bf{ens.}}} \sum_{n=1}^{N_{\bf{ens.}}} \operatorname{Var} \Big( Z_n^{\theta}(s,a) \mid Z_n^{\theta}(s,a) \le \operatorname{VaR}_\beta(Z_n^{\theta}) \Big) }.
\end{aligned}
\end{equation}

The total variance equals the sum of epistemic variance and aleatoric variance. Consequently, the total uncertainty is formulated as $\sigma_{\text{total}} = \sqrt{\sigma_{\text{epi.}}^2 + \sigma_{\text{ale.}}^2}$.

\subsection{Uncertainty-embedded HOCBF}
\label{sec:ue_cons}
\subsubsection{Safety Correction Formulation}
The learned risk-sensitive policy $\boldsymbol{u}^{\text{RL}}$ still lacks formal safety guarantees. Particularly in scenarios where the policy exhibits low confidence, its unreliable decisions may introduce critical risks. We therefore introduce CBF as a safety filter to  enhance safety.

A key challenge in applying standard CBFs to autonomous driving is that the control input $\boldsymbol{u}$ must appear explicitly in the first derivative of the barrier function $h(\boldsymbol{x})$. However, when safety constraints involve inter-vehicle distances but the control inputs are vehicle accelerations, this assumption is frequently violated. This operational requirement is formalized by the notion of relative degree.

\begin{definition} [Relative Degree]
A continuously differentiable function $h:\mathcal{X}\rightarrow\mathbb{R}$ is said to have a relative degree $r\in\mathbb{N}$ with respect to the system (\ref{eq:con_affine_sys}) if, for all $\boldsymbol{x}\in\mathcal{X}$, $L_gL_f^ih(\boldsymbol{x})=0$ for all $i\in\{0,1,\dots,r-2\}$, and $L_gL_f^{r-1}h(\boldsymbol{x})\neq 0$.
\end{definition}

To address this, High-Order CBFs recursively define a sequence of auxiliary functions starting with $\Psi_0(\boldsymbol{x}) := h(\boldsymbol{x})$:
\begin{equation}
\begin{aligned}[b]
\Psi_i(\boldsymbol{x}) := \dot{\Psi}_{i-1}(\boldsymbol{x}) + \alpha_i(\Psi_{i-1}(\boldsymbol{x})), \quad \forall i \in \{1, \dots, r\},
\end{aligned}
\end{equation}
where each $\alpha_i$ is a differentiable class $\mathcal{K}$ function. Then, a series of superlevel set $\mathcal{C}_i$ can be represented as:
\begin{equation}
\begin{aligned}[b]
\mathcal{C}_i:=\left\{x\in\mathcal{X}:\Psi_{i-1}(x)\geq0\right\},\quad\forall i\in\{1,\ldots,r\}.
\label{E13}
\end{aligned}
\end{equation}

\begin{definition} [High-Order CBF \cite{xiong2023dtcbf}]
Given a sequence of sets $\mathcal{C}_i$ for $i\in\{1,\dots,r\}$ as in (\ref{E13}) and the auxiliary functions $\Psi_i(\boldsymbol{x})$ defined recursively, a continuously differentiable function $h:\mathcal{X}\rightarrow\mathbb{R}$ is a valid HOCBF of relative degree $r$ if there exist class $\mathcal{K}$ functions $\alpha_i$ ($\forall i\in\{1,\ldots,r\}$) such that:
\begin{equation}
\begin{aligned}[b]
\sup_{\boldsymbol{u} \in \mathcal{U}} \Psi_r(\boldsymbol{x},\boldsymbol{u}) \geq 0,\quad\forall \boldsymbol{x} \in \bigcap\limits_{i=1}^{r} {\mathcal{C}_i}.
\label{E14}
\end{aligned}
\end{equation}
\end{definition}

Explicitly, expanding the condition in (\ref{E14}) using Lie derivatives yields the forward invariance constraint:
\begin{equation}
\begin{aligned}
\sup_{\boldsymbol{u} \in \mathcal{U}} \Bigl[
& L_f^r h(\boldsymbol{x})
+ L_g L_f^{r-1} h(\boldsymbol{x}) \boldsymbol{u} \\
& + O(h(\boldsymbol{x}))
+ \alpha_r(\Psi_{r-1}(\boldsymbol{x}))
\Bigr]
\ge 0
\end{aligned}
\end{equation}
where $O(h(\boldsymbol{x}))$ denotes the remaining scalar terms accumulated from the recursive Lie derivatives of the lower-order auxiliary functions and the associated class $\mathcal{K}$ functions. 

\begin{figure}
    \centering
    \subfloat{%
        \includegraphics[width=0.35\textwidth]{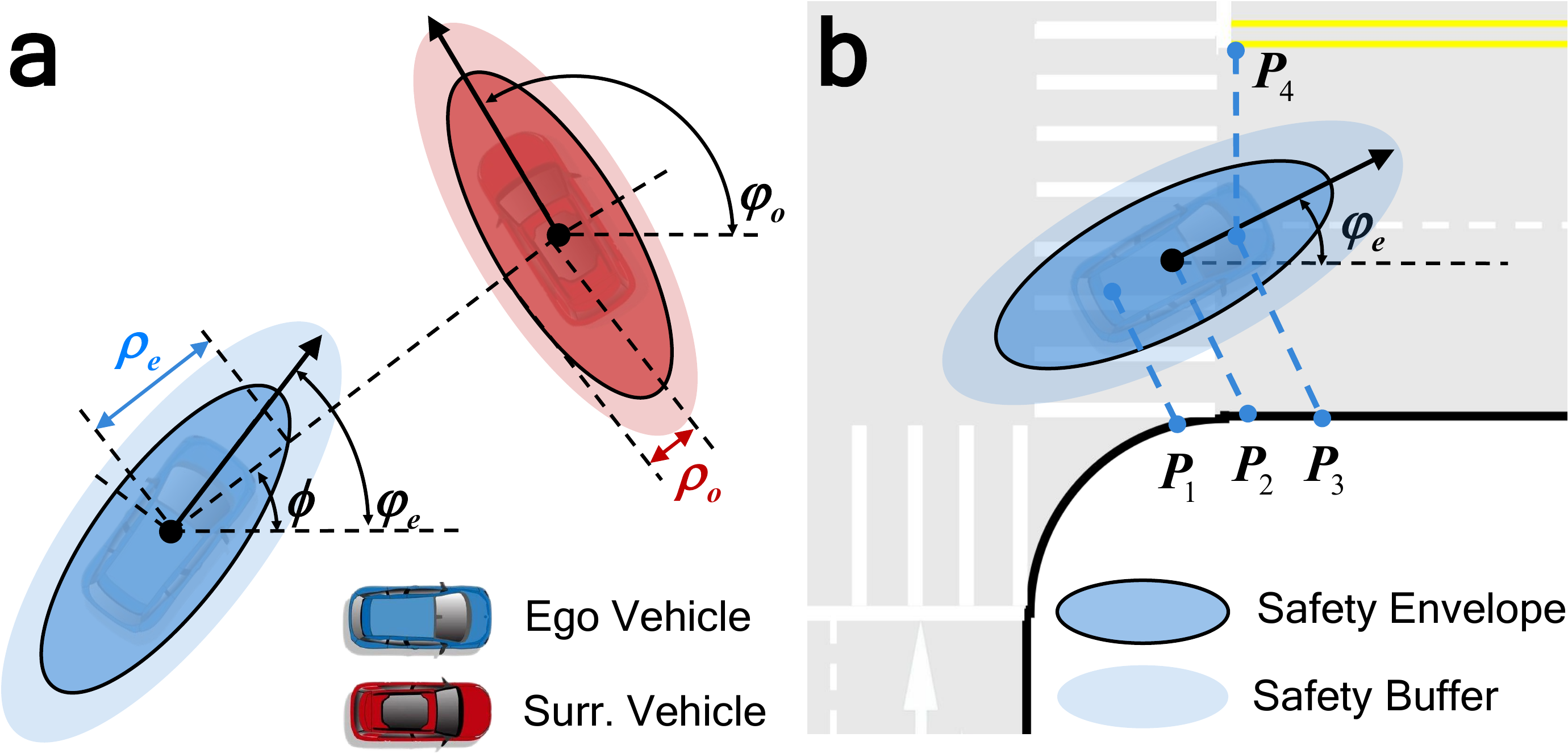}
        \label{safety_constarints}
    }
    \caption{Spatial safety constraints. Elliptical envelopes are utilized to describe the spatial constraints among the EV, SVs, and road boundaries. a) Directional support radii $\rho_e, \rho_o$ are calculated using vehicle headings $\varphi_e, \varphi_o$ and the relative angle $\phi$, with an added safety buffer $d_{safe}$. b) To establish road boundary constraints, the EV's center of gravity and wheelbase-based reference points (at 1/3 and 2/3 intervals) are projected onto the lane edges.}
    \label{fig:safety_constraints}
\end{figure}

To implement this theoretical framework for our EV, we construct a control-affine system based on the single-track kinematic model~\cite{chourasiya2026input}. We define the EV's state vector as $\boldsymbol {x} \triangleq [p_x,\;p_y,\;\psi,\;v]^\top$ and the control input as $\boldsymbol {u} \triangleq [a_{\text{lon}},\;z_{f}]^\top$. Here, a pseudo-input $z_f=\tan\delta_f$ replaces the direct steering angle to satisfy the control-affine structure. As illustrated in Fig.~\ref{fig:safety_constraints}, the spatial constraints imposed by surrounding vehicles and road boundaries are concurrently considered. For an arbitrary obstacle $i$, let its center position be $(o_x^i, o_y^i)$, velocity be $(v_x^i, v_y^i)$, and heading be $\psi_i$. We define an augmented state vector $\boldsymbol{x}_i \triangleq [p_x, p_y, \psi, v, o_x^i, o_y^i]^\top$ to establish the relative kinematics. Assuming obstacles maintain constant velocity within the prediction horizon, the augmented control-affine dynamics are given by:
\begin{equation}
\begin{aligned}[b]
\dot{\boldsymbol{x}}_i =
\begin{bmatrix}
v\cos\psi \\
v\sin\psi \\
0 \\
0 \\
v_x^i \\
v_y^i
\end{bmatrix}
+
\begin{bmatrix}
0 & 0 \\
0 & 0 \\
0 & \frac{v}{L} \\
1 & 0 \\
0 & 0 \\
0 & 0
\end{bmatrix}
\boldsymbol{u}.
\end{aligned}
\end{equation} where $L$ denotes the wheelbase of the EV. Furthermore, we define the relative position vector $r_i$ and the relative angle $\phi_i$ as:
\begin{equation}
\begin{aligned}[b]
r_i \triangleq
\begin{bmatrix}
p_x - o_x^i \\
p_y - o_y^i
\end{bmatrix}, \qquad
\phi_i \triangleq \operatorname{atan2}(p_y - o_y^i, p_x - o_x^i).
\end{aligned}
\end{equation}

To unify the representations of heterogeneous obstacles, we introduce a boolean mode indicator $m_i \in \{0, 1\}$, where $m_i = 1$ corresponds to dynamic vehicle obstacles and $m_i = 0$ corresponds to static road boundary points. The directional support radii for the EV and the obstacle are derived via an elliptical approximation:
\begin{equation}
\begin{aligned}[b]
\rho_e^i &\triangleq \sqrt{a_e^2\cos^2(\phi_i - \psi) + b_e^2\sin^2(\phi_i - \psi) + \varepsilon_0}, \\
\rho_o^i &\triangleq \sqrt{(a_o^i)^2\cos^2(\phi_i - \psi_i) + (b_o^i)^2\sin^2(\phi_i - \psi_i) + \varepsilon_0},
\end{aligned}
\end{equation} where $\varepsilon_0 > 0$ is a small constant for numerical regularization. The effective safety margin distance is then synthesized as:
\begin{equation}
\begin{aligned}[b]
d_i \triangleq m_i(\rho_e^i + \rho_o^i) + (1 - m_i)\bar{d}_i +d_{\text{safe}},
\end{aligned}
\end{equation} where $\bar{d}_i$ is the static margin for road boundaries, $d_{\text{safe}}$ is the safety buffer. Consequently, the candidate barrier function is formulated as:
\begin{equation}
\begin{aligned}[b]
h_i(\boldsymbol{x}_i) \triangleq \|r_i\|_2^2 - d_i^2.
\end{aligned}
\end{equation}

The forward invariant safe set is defined as $\mathcal{C} \triangleq \bigcap_{i\in\mathcal{I}} \{\boldsymbol{x}_i \mid h_i(\boldsymbol{x}_i) \ge 0\}$, where $\mathcal{I} = \mathcal{I}_{\text{veh}} \cup \mathcal{I}_{\text{road}}$ encompasses all surrounding entities. Since the control input $\boldsymbol{u}$ first appears in the second derivative of $h_i$ (i.e., relative degree is 2), a generalized HOCBF constraint for each obstacle is expressed using Lie derivatives:
\begin{equation}
\begin{aligned}[b]
\mathcal{H}_i(\boldsymbol{x}_i, \boldsymbol{u}) \triangleq \sup_{\boldsymbol{u} \in \mathcal{U}} \Bigl[
L_f^2 h_i(\boldsymbol{x}_i) + L_g L_f h_i(\boldsymbol{x}_i)\boldsymbol{u} + \\O(h_i(\boldsymbol{x}_i)) + \alpha_2(\Psi_1(\boldsymbol{x}_i))\Bigl]\ge 0,
\end{aligned}
\end{equation}

At each sampling instant, given the nominal RL policy action $\boldsymbol{u}^{\text{RL}}$, the safety filter computes the minimally invasive safe action by solving the following Quadratic Program (QP):
\begin{equation}
\begin{aligned}[b]
\min_{\boldsymbol{u}} \quad & \frac{1}{2}(\boldsymbol{u} - \boldsymbol{u}^{\text{RL}})^\top W_u (\boldsymbol{u} - \boldsymbol{u}^{\text{RL}}) \\
\text{s.t.} \quad & a_{\min} \le a \le a_{\max}, \quad z_{\min} \le z \le z_{\max}, \\
& \mathcal{H}_i(\boldsymbol{x}_i, \boldsymbol{u})\ge 0, \quad \forall i \in \mathcal{I}_{\text{veh}} \cup \mathcal{I}_{\text{road}}, \\
\end{aligned}
\end{equation} Here, $W_u = \operatorname{diag}(w_a, w_z)$ is the control weight matrix. Finally, the actual steering command applied to the vehicle is recovered via $\delta_f^\star = \arctan(z^\star)$. The details of specific HOCBFs can be seen in Appx.~\ref{apx:hocbf}.

\subsubsection{Uncertainty-embedded Constraints}
We posit that AVs should execute safe, conservative maneuvers under high uncertainty while leveraging greater flexibility under low uncertainty, rather than strictly adhering to conservative policies that inevitably degrade traffic efficiency. Consider a nonlinear control-affine system subject to uncertainties, governed by $\dot{\boldsymbol{x}}=\hat{f}(\boldsymbol{x})+\hat{g}(\boldsymbol{x})\boldsymbol{u}+\varphi(\boldsymbol{x},\boldsymbol{u})$, where $\hat{f}$ and $\hat{g}$ denote the known nominal dynamics, and $\varphi(\boldsymbol{x},\boldsymbol{u})$ represents the unmodeled dynamic interactions. By embedding this uncertainty into a  CBF, the safety condition for all states $\boldsymbol{x} \in \mathcal{C}$ is formulated as:
\begin{equation}
\sup_{\boldsymbol{u}\in\mathcal{U}}\inf_{w\in\mathcal{W}}\left[L_{\hat{f}}h(\boldsymbol{x})+L_{\hat{g}}h(\boldsymbol{x})\boldsymbol{u}+w(\boldsymbol{x},\boldsymbol{u})\right]\geq-\alpha(h(\boldsymbol{x}))
\end{equation} where $w(\boldsymbol{x},\boldsymbol{u})=L_{\varphi}h(\boldsymbol{x})$ is the scalar projection of the uncertainty onto the gradient of $h$. 

In the discrete-time domain, we treat the CBF as a one-step predictor: $\hat h_{k+1|k}=h(\boldsymbol{x}_k)+\Delta h(\boldsymbol{x},\boldsymbol{u})$. The true next-step barrier value is $h(\boldsymbol{x}_{k+1})=\hat h_{k+1|k}+w_{k}$, where the residual $w_{k}$ captures the model mismatch. To estimate the residual online, we train an MLP predictor with two separate head for vehicle and road constraints. Given the feature vector $\xi_{i,k}$ composed of the normalized RL action, ego state, and obstacle state, the network predicts the residual $\hat{w}_k = f_{\epsilon,c}(\xi_{k})$, where $c\in\{\mathrm{vehicle},\mathrm{road}\}$ denotes the obstacle type, and $\xi_k$ represents the feature vector.

Crucially, any negative residual ($\hat{w}_k < 0$) indicates that the nominal model overestimates safety, necessitating a compensatory tightening of the constraint. Meanwhile, when the RL policy exhibits high confidence (low $\sigma_k$), the constraint can be appropriately relaxed. Consequently, the uncertainty-embedded CBF is formulated as:
\begin{equation}
\sup_{\boldsymbol{u}\in\mathcal{U}}\left[\Delta h(\boldsymbol{x},\boldsymbol{u}) + \hat{w}_{k} + \delta(\sigma_k) + \alpha(h(\boldsymbol{x}))\right]\geq 0
\end{equation}
where the adaptive slack term $\delta(\sigma_k)$ is a piecewise continuous and strictly decreasing function anchored by uncertainty quantiles:
\begin{equation}
\delta(\sigma_k) = \begin{cases} \eta(\sigma_{\ell} - \sigma_k), & \sigma_k < \sigma_h \\ \eta(\sigma_{\ell} - \sigma_h) - 2\eta(\sigma_k - \sigma_h), & \sigma_k \ge \sigma_h \end{cases}
\end{equation} where $\sigma_{\ell}$ and $\sigma_{h}$ represent the 95th and 97.5th percentiles of the long-tailed uncertainty distribution. $\eta > 0$ is the modulation gain. This formulation ensures $\delta(\sigma_k)$ remains positive for $\sigma_k < \sigma_{\ell}$, promoting efficiency through relaxation, while doubling the tightening slope beyond $\sigma_h$ to enforce a rigorous safety margin against long-tail risks.

\section{Implementation}
\subsection{Simulation Environment}
We implement a bidirectional four-lane intersection scenario based on Highway-Env \cite{highwayenv}. Each SV is controlled by an improved Intelligent Driver Model (IDM) \cite{idm}, which predicts its heading and position for the subsequent 2 s, yielding to potential collisions according to road priorities. During training, the EV learns from a mixed distribution of tasks to build a generalizable policy. To comprehensively validate the framework, the testing phase is partitioned into two regimes with distinct evaluation focuses. The Random Destination task assesses general navigation capabilities under a standard traffic density (e.g., 20 veh/km). Conversely, the Dense Unprotected Left Turn task establishes a stress test by significantly elevating traffic density (e.g., 30 veh/km) and maximizing conflict points. This surge in density fundamentally compresses the distribution of acceptable time headways, statistically increasing the probability of encountering OOD and long-tail events. When resetting the scenarios, SVs are initialized with random velocities between 6 m/s and 10 m/s, and the EV is placed in a collision-free lane with a random velocity. Simulation frequency $f_{s}$ is 15 Hz, with the policy execution frequency $f_{\pi}$ set to 10 Hz.

\subsection{MDP Formulation}
\subsubsection{Observation and Action Spaces}
The observation consists of the state of EV $\mathcal{S}_{EV}$, the states of $N_{SV}$ SVs $\mathcal{S}_{SV}$, the next $N_{wp}$ reference waypoints $\mathcal{S}_{wp}$, and a one-hot task encoding $\mathcal{S}_{task}$ for left turn, going straight, and right turn: $\mathcal{S}=[\mathcal{S}_{EV},\mathcal{S}_{SV},\mathcal{S}_{wp},\mathcal{S}_{task}].$
The ego state is defined as $\mathcal{S}_{EV}=[\mathbb{I}_{veh},x,y,v_x,v_y,\varphi,\omega,d_{veh},d_{road},d_{des}],
$
where $x$ and $y$ are the coordinates of the vehicle center of gravity, $v_x$ and $v_y$ are the longitudinal and lateral velocities, $\varphi$ is the heading angle, and $\omega$ is the yaw rate. Here, $\mathbb{I}_{veh}\in\{0,1\}$ is a validity indicator, with $\mathbb{I}_{veh}=1$ for the ego vehicle. The quantities $d_{veh}$, $d_{road}$, and $d_{des}$ denote the distances to the nearest vehicle, the road boundary, and the destination, respectively. The surrounding-vehicle state is $\mathcal{S}_{SV}=\{\mathcal{S}^{j}_{SV}\}_{j=1}^{N_{SV}},
\mathcal{S}^{j}_{SV}=[\mathbb{I}_{veh},\Delta x,\Delta y,\Delta v_x,\Delta v_y,\Delta\varphi],$
where $\Delta x$, $\Delta y$, $\Delta v_x$, $\Delta v_y$, and $\Delta\varphi$ are the position, velocity, and heading of the $j$-th surrounding vehicle relative to the ego vehicle. The waypoints in $\mathcal{S}_{wp}$ are represented by their relative position offsets with respect to the ego vehicle. The continuous action space is defined as $\mathcal{A}=[a_{\mathrm{lon}},\delta_f],$ where $a_{\mathrm{lon}}$ and $\delta_f$ denote the longitudinal acceleration and front-wheel steering angle, respectively.

\subsubsection{Reward Design}
The total reward is composed of sparse and dense terms: $\mathbf{r}=\mathbf{r}_{sparse}+\mathbf{r}_{dense}.$ The sparse reward penalizes collisions and rewards successful task completion:
\begin{subequations}
\label{eq:reward_sparse}
\begin{align}
&\mathbf{r}_{sparse} = \mathbf{r}_{collision}+\mathbf{r}_{arrive\_goal}, \\
&\mathbf{r}_{collision} = -50\,\mathbb{I}_{collision}, \\
&\mathbf{r}_{arrive\_goal} = 50\,\mathbb{I}_{arrive\_goal}.
\end{align}
\end{subequations}

The dense reward accounts for reference-line tracking, action smoothness, progress to the destination, and safety:
\begin{subequations}
\label{eq:reward_dense}
\begin{align}
&\mathbf{r}_{dense} = \frac{3}{1+\mathbf{r}_{ref}}+\mathbf{r}_{act}+\mathbf{r}_{des}+\mathbf{r}_{safe}, \\
&\mathbf{r}_{ref} = \min_{i\in\{1,2\}}
(\boldsymbol{x}_{k,i}^{\mathrm{ref}}-\boldsymbol{x}_k)^\top
Q(\boldsymbol{x}_{k,i}^{\mathrm{ref}}-\boldsymbol{x}_k), \\
&\mathbf{r}_{act} = -\left(\boldsymbol{u}_k^\top R_a \boldsymbol{u}_k
+ \Delta \boldsymbol{u}_k^\top R_{\Delta}\Delta \boldsymbol{u}_k\right), \\
&\mathbf{r}_{des} = -0.02\cdot d_{des}^2, \\
&\mathbf{r}_{safe} =
\begin{cases}
0, & d_{veh}>2.0, \\
-(2.0-d_{veh}), & 0.5<d_{veh}\le 2.0, \\
-3(1.0-d_{veh}), & d_{veh}\le 0.5.
\end{cases}
\end{align}
\end{subequations} Here, $\boldsymbol{x}_{k,i}^{\mathrm{ref}}=
[x_i^{\mathrm{ref}},y_i^{\mathrm{ref}},v_{x,i}^{\mathrm{ref}},0,\varphi_i^{\mathrm{ref}},0]^\top.$ For $\mathbf{r}_{ref}$, the smaller tracking error among the two candidate reference lines is selected, encouraging the ego vehicle to follow the more suitable path. The term $\mathbf{r}_{act}$ penalizes large control inputs and abrupt action variations, thereby improving smoothness and energy efficiency. The weight matrices are
$ Q=\mathrm{diag}(400.0,400.0,30.0,30.0,2.0,0.5), R_a=\mathrm{diag}(0.05,0.02), R_{\Delta}=\mathrm{diag}(0.2,0.3). $

\subsection{Network Architecture and Training Details}
\begin{figure*}[ht]
    \centering
    \subfloat{%
        \includegraphics[width=0.95\textwidth]{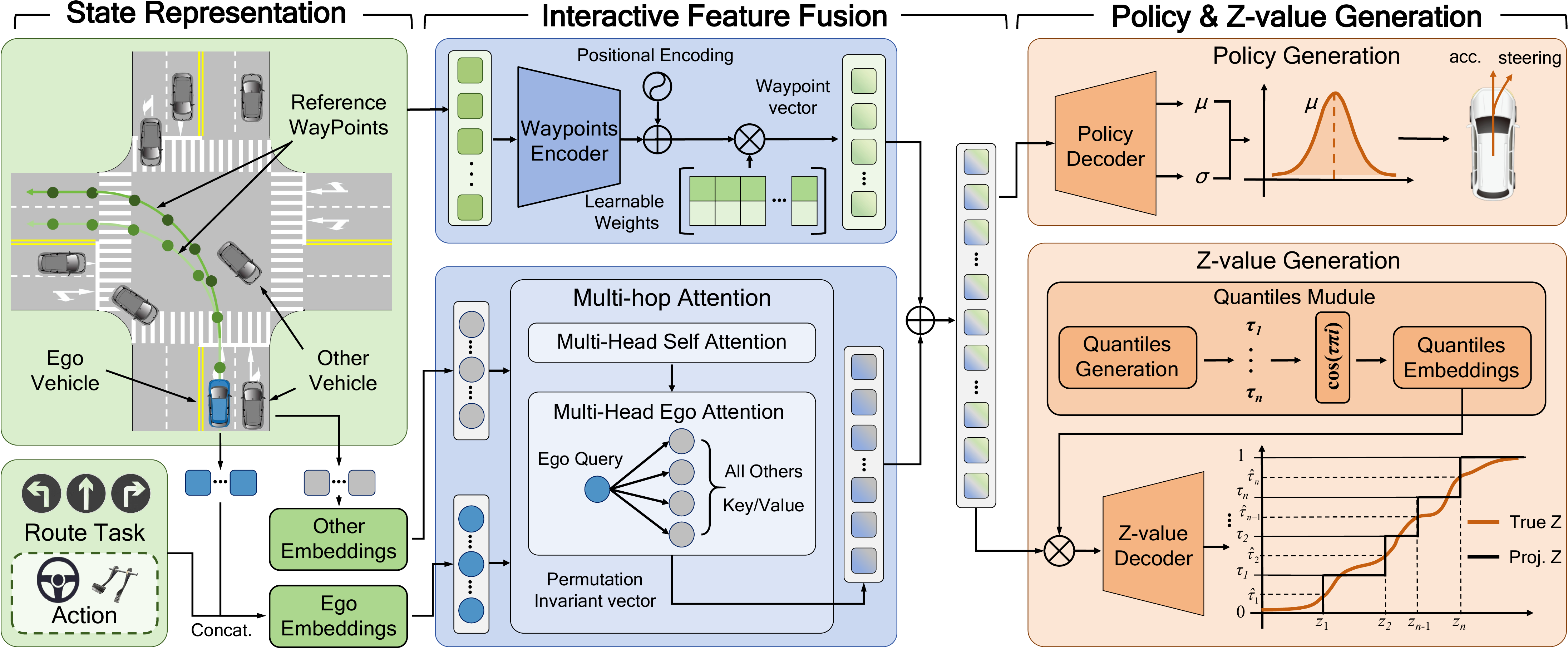}
    }
    \caption{Network architecture. The state components $\mathcal{S}_{\text{EV}}$, $\mathcal{S}_{\text{SV}}$, and $\mathcal{S}_{\text{wp}}$ are individually encoded by dedicated embedding layers into latent representations. An interactive feature fusion module then integrates a positional-encoded waypoint branch with learnable weights and a multi-hop attention branch, which extracts permutation-invariant interaction features between the ego vehicle and surrounding traffic. The resulting fused latent representation is subsequently fed into two parallel decoders: one generates the policy distribution, and the other produces the quantile-based $Z$-value estimates for distributional critic learning.}
    \label{fig:network_architechture}
\end{figure*}
Building upon our previous work \cite{leng2025risk}, we adopt the network architecture illustrated in Fig.~\ref{fig:network_architechture}, which is specifically designed to accommodate the dynamic nature of unsignalized intersection scenarios, i.e., varying numbers of surrounding vehicles and their constantly changing spatial configurations. It should be emphasized that the primary focus of this paper is not the specific method for generating quantiles; accordingly, while fixed quantile fractions are employed in the reported experiments for simplicity, the proposed framework does not preclude the adoption of more advanced or adaptive quantile-generation techniques in future extensions. Regarding the training of the CBF residual predictor, the required dataset is collected prior to RL policy learning by executing random actions within the simulation environment. The predictor is trained to regress the residual $w_k = h(\boldsymbol{x}_{k+1}) - \hat{h}_{k+1|k}$, where $\hat{h}_{k+1|k}$ denotes the nominal one-step ahead prediction of the barrier function and $h(\boldsymbol{x}_{k+1})$ is its true observed value. The detailed regression performance of the residual predictor is provided in Appx.~\ref{apx:cbf_predictor}. The complete set of hyperparameters utilized throughout this work is summarized in Tab.~\ref{hyper}.

\begin{table}
    \setlength\tabcolsep{1pt}
    \centering
    \caption{Hyper-Parameters}
    \scriptsize
    \begin{tabular}{lrlr}
        \toprule 
         Hyper-parameter & Value & Hyper-parameter & Value\\ 
        \midrule 
        Network hidden size &  256 & Temperature factor & 0.005\\
        Activation function & GELU & Batch size & 256\\
        Actor learning rate $\alpha_{\pi}$ & 3e-4→1e-5& Entropy learning rate & 3e-4\\
        Critic learning rate $\alpha_{r,c}$ & 3e-3→1e-4 & Target entropy $\bar{\mathcal{H}}$ & -\text{dim}($\mathcal{A}$)\\
        Discount factor $\gamma$ & 0.99 & Number of ensemble $N$ & 5\\
        Prior factor $\rho$ & 10.0 & Bernoulli mean $\mathrm{p}$ & 0.9\\
        Learning buffer size & 1e5 & Number of quantile sample $N_q$ & 32\\
        \text{CVaR} risk parameter $\beta$ & 0.25 & CBF residual dataset size & 2e5\\
        Hidden size of predictor $f_{\epsilon,c}$ & 64 & Learning rate of predictor $f_{\epsilon,c}$ & 3e-4\\ 
        Training epoch of predictor $f_{\epsilon,c}$ & 300 & Batch size of predictor $f_{\epsilon,c}$ & 4096 \\
        TTCBF class $\mathcal{K}$ function $\lambda_1^T$ & 0.1 & TTCBF parameter $\Gamma$ & 100 \\
        ECBF class $\mathcal{K}$ function $\lambda_1^E$ & 1.5 &  ECBF class $\mathcal{K}$ function $\lambda_2^E$ & 0.3 \\
        \bottomrule
    \end{tabular}
    \label{hyper}
\end{table}

\subsection{Baselines and Evaluation Metrics}
\begin{table}
    \scriptsize
    \centering
    \rowcolors{2}{white}{gray!20}
    \caption{Evaluation Metrics of Decision and Control Algorithm}
    \begin{tabularx}{0.48\textwidth}{>{\raggedright\arraybackslash}p{2.5cm}X}
    \toprule
    \textbf{Metric name}       & \textbf{Description}\\
    \midrule
Success Rate (SR)        & Percentage of episodes successfully reaching the target without violations or timeouts. \\
Frozen Rate (FR)         & Percentage of episodes timing out due to stagnation or insufficient progress. \\
Route Completion (RC)    & Ratio of traveled distance along the reference trajectory to the planned route length. \\
Average Episode Speed (AES) & Average speed of the ego vehicle during intersection navigation. \\
Average Episode Reward (AER)   & Cumulative reward obtained by the agent in a single episode. \\
\midrule
Violation Rate (VR)      & Percentage of episodes ending in safety-critical failures like collisions or traffic rule infractions. \\
Min Distance to Closest Vehicle (min-DTC) & Minimum gap to the nearest obstacle throughout an episode, indicating the safety margin. \\
\bottomrule
\end{tabularx}
\label{tab:metrics}
\end{table}

To verify the generalizability of the proposed RUDC framework, we integrate it with two distinct High-Order Control Barrier Function (HOCBF) algorithms: Exponential CBFs (ECBFs)~\cite{7524935} and Truncated Taylor CBFs (TTCBFs)~\cite{xu2025high}, resulting in two variants denoted as RUDC-E and RUDC-T, respectively. We then compare our algorithms against several baseline methods: Quadratic Programming Safety Layer (QPSL)~\cite{dalal2018safe}, Recovery RL~\cite{thananjeyan2021recovery}, Feasible Actor Critic (FAC)~\cite{ma2021feasible}, Unrolling Safety Layer (USL)~\cite{zhang2023evaluating}, Lagrangian Relaxation~\cite{lagrangian}, vanilla SAC~\cite{huber1992robust} and distributional SAC (DSAC)~\cite{ma2025dsac}. To ensure a fair comparison, all baselines employ network architectures comparable to those adopted in this work, with the sole distinction that their critic components are limited to standard value estimation. Detailed implementation settings for all algorithms are provided in Appx.~\ref{apx:saferl}. To comprehensively evaluate the performance of the EV, we adopt the metrics listed in Tab.~\ref{tab:metrics}.

\section{Experiment Results}
\subsection{Comparison with Baselines}
\begin{figure*}
    \centering
    \subfloat{%
        \includegraphics[width=0.70\textwidth]{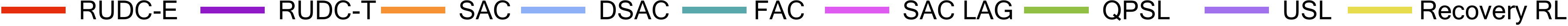}
    }\\ \vspace{-5pt}
    \subfloat{%
        \includegraphics[width=0.24\textwidth]{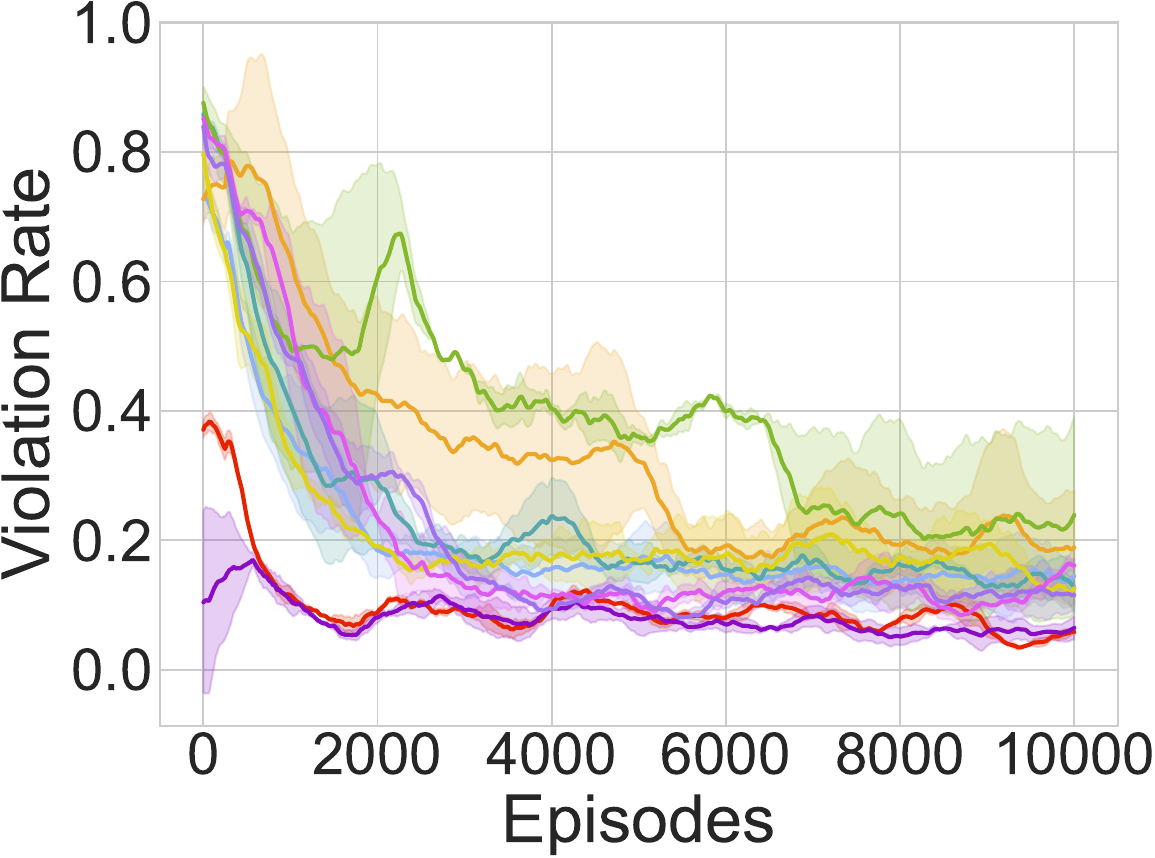}
        \label{compare_violate}
    }\subfloat{%
        \includegraphics[width=0.24\textwidth]{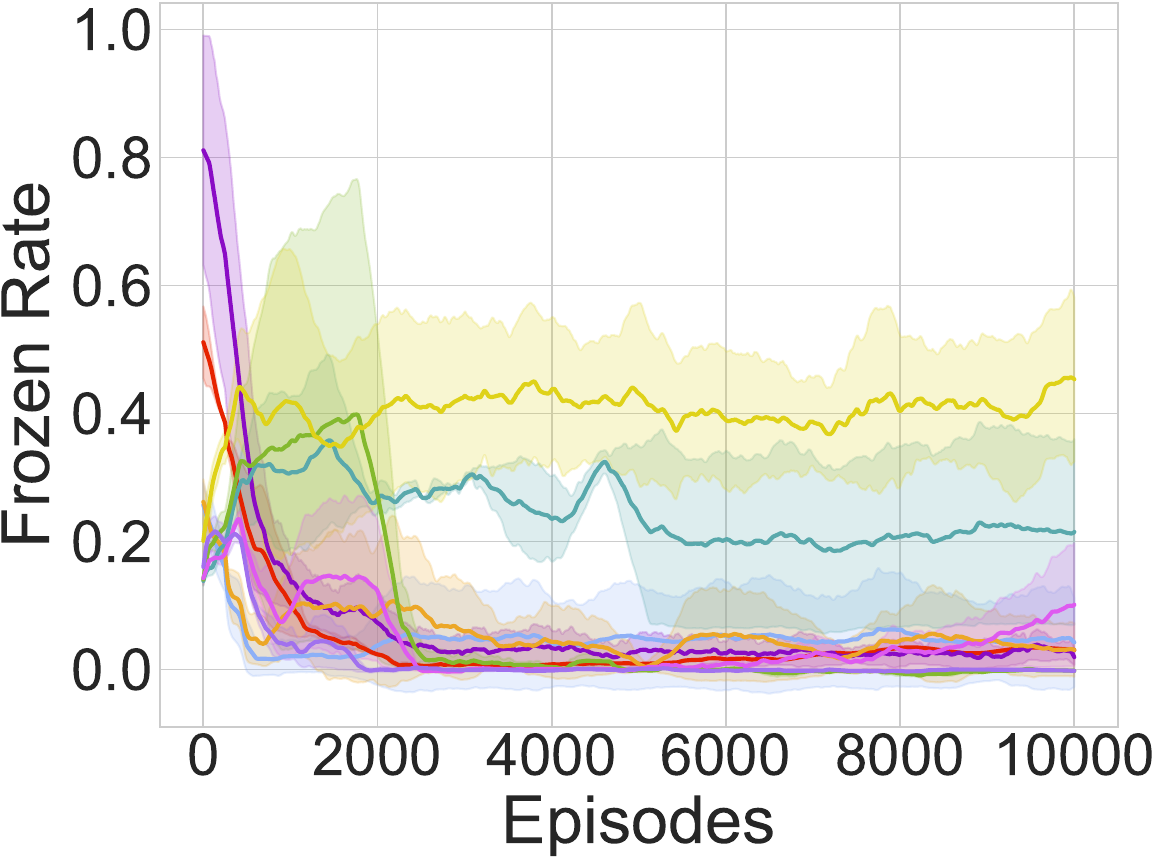}
        \label{compare_frozen}
    }\subfloat{%
        \includegraphics[width=0.24\textwidth]{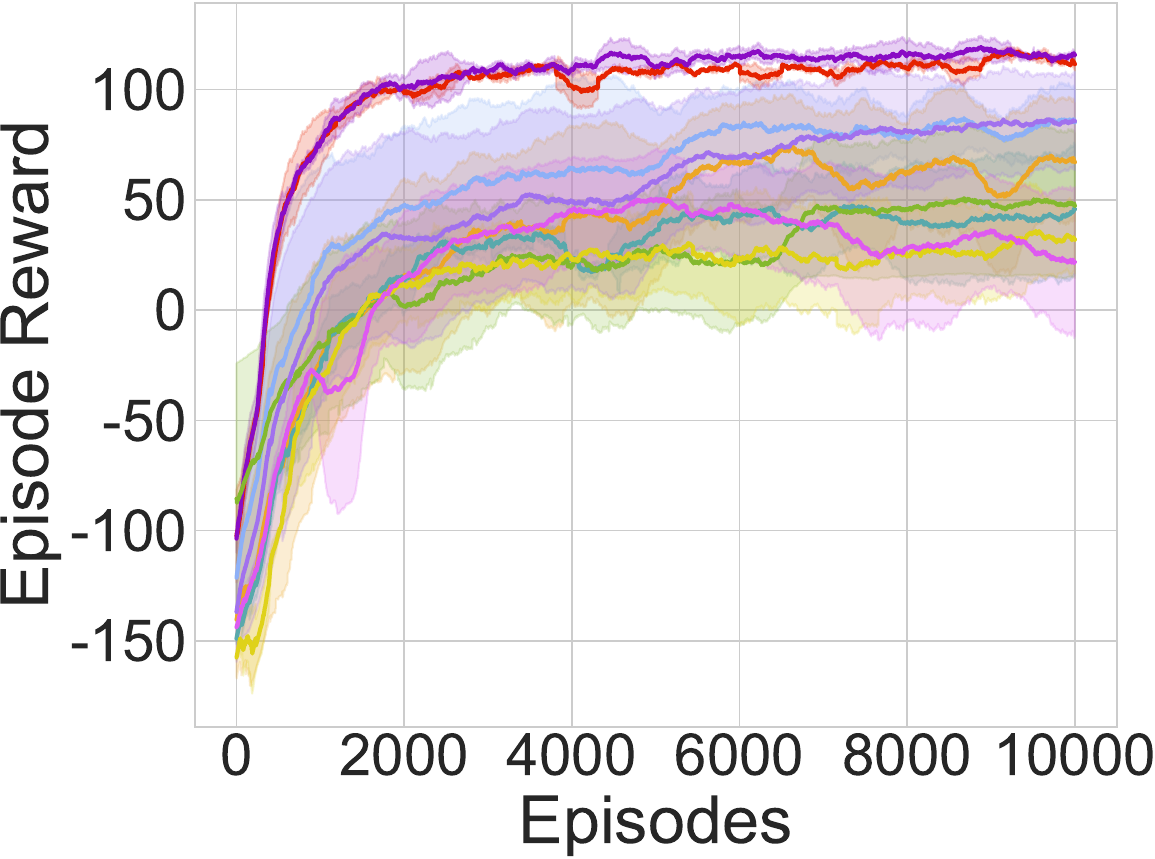}
        \label{compare_reward}
    }\subfloat{%
        \includegraphics[width=0.24\textwidth]{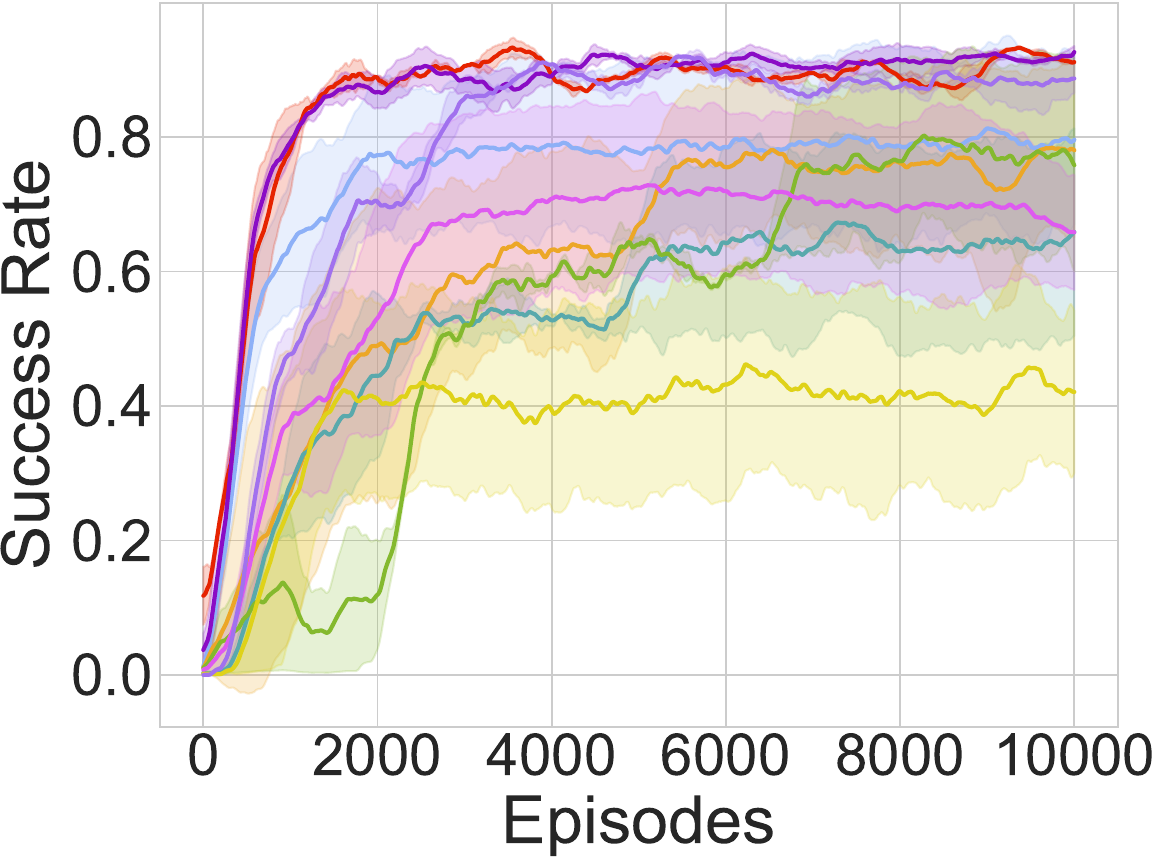}
        \label{compare_success}
    }
    \caption{Training curves for comparison experiments. Solid lines denote means, while shaded regions represent 95\% confidence intervals over three runs.}
    \label{fig:compare}
\end{figure*}

\begin{table*}
    \centering
    \scriptsize
    \caption{Performance comparison across two Test Conditions.
    $\uparrow$\,/\,$\downarrow$ indicates higher\,/\,lower is better.
    \textbf{Bold} and \underline{underlined} denote the best and second-best results, respectively. Values are presented as mean (standard deviation).}
    \label{tab:comparison}
    \setlength{\tabcolsep}{2.4pt}
    \renewcommand{\arraystretch}{1.25}
    \begin{tabular}{l c c c c c >{\columncolor{gray!15}}c >{\columncolor{gray!15}}c c c c c c >{\columncolor{gray!15}}c >{\columncolor{gray!15}}c}
        \toprule
        \textbf{Algorithm}
        & \multicolumn{7}{c}{\textit{Random Destination}}
        & \multicolumn{7}{c}{\textit{Dense Unprotected Left Turn}} \\
        \cmidrule(lr){2-8} \cmidrule(lr){9-15}
        & SR\,$\uparrow$ & FR\,$\downarrow$ & RC\,$\uparrow$ & AES\,$\uparrow$ & AER\,$\uparrow$ & VR\,$\downarrow$ & min-DTC\,$\uparrow$
        & SR\,$\uparrow$ & FR\,$\downarrow$ & RC\,$\uparrow$ & AES\,$\uparrow$ & AER\,$\uparrow$ & VR\,$\downarrow$ & min-DTC\,$\uparrow$ \\
        & (\%) & (\%) & (\%) & (m/s) & & (\%) & (m)
        & (\%) & (\%) & (\%) & (m/s) & & (\%) & (m) \\
        \midrule
        QPSL
        & 77.3(15.0) & \underline{0.2(0.1)} & 95.7(7.2) & \textbf{8.9(1.1)} & 47.8(34.1) & 22.5(14.5) & 2.9(3.3)
        & 63.3(21.3) & \textbf{0.6(0.3)} & 90.1(14.5) & \underline{8.4(0.6)} & 11.2(51.1) & 36.1(19.8) & 2.0(3.2) \\

        Recovery RL
        & 43.0(14.3) & 42.0(12.4) & 91.3(8.8) & 6.5(2.2) & 31.6(21.6) & 15.0(1.9) & 5.5(4.7)
        & 36.9(19.1) & 47.3(16.6) & 85.8(14.6) & 5.4(3.6) & -40.6(71.6) & 15.8(4.1) & \textbf{6.7(5.5)} \\

        FAC
        & 64.4(15.7) & 22.1(15.8) & 90.1(9.9) & 6.6(2.4) & 41.8(29.6) & 13.6(2.6) & 2.2(2.7)
        & 56.1(17.8) & 21.6(10.2) & 87.7(16.1) & 7.1(1.4) & 7.2(45.0) & 22.3(7.6) & 1.2(1.7) \\

        USL
        & 88.5(3.7) & \textbf{0.0(0.1)} & 96.6(6.4) & \underline{8.7(1.4)} & 84.6(23.7) & 11.5(3.8) & 4.8(3.7)
        & 81.6(8.2) & \underline{1.9(0.5)} & 92.3(13.0) & \underline{8.4(1.6)} & 49.6(39.1) & 16.5(6.3) & 1.6(2.1) \\

        Lagrangian
        & 74.5(10.5) & 13.1(7.4) & 95.3(8.7) & 7.6(1.2) & 28.1(28.8) & 12.4(1.4) & 4.5(3.8)
        & 67.8(14.1) & 9.4(9.4) & 93.6(12.3) & 8.1(1.1) & 1.3(47.5) & 23.1(3.8) & 2.3(1.9) \\

        SAC
        & 75.4(15.3) & 3.5(4.6) & 94.4(7.6) & 8.5(1.7) & 61.1(32.3) & 21.1(10.7) & 2.1(3.6)
        & 61.3(19.1) & 3.3(3.9) & 81.6(11.2) & \textbf{8.7(1.4)} & -51.9(82.3) & 35.3(15.8) & 1.1(1.9) \\

        DSAC
        & 80.8(14.5) & 4.6(7.6) & 95.1(7.1) & 8.3(2.1) & 81.9(14.3) & 14.6(5.3) & 3.4(3.1)
        & 71.3(18.2) & 7.1(3.9) & 89.6(10.5) & 7.7(1.8) & 31.2(35.7) & 21.6(8.1) & 1.7(2.3) \\

        \midrule
        \textbf{RUDC-T}
        & \textbf{92.3(1.2)} & 3.2(1.1) & \textbf{98.5(5.9)} & 7.3(1.3) & \textbf{115.0(4.6)} & \textbf{4.5(0.2)} & \textbf{6.3(4.3)}
        & \underline{86.3(2.1)} & 5.8(1.2) & \textbf{97.6(6.1)} & 6.7(2.2) & \textbf{83.0(16.2)} & \textbf{7.9(1.6)} & \underline{5.7(4.9)} \\

        \textbf{RUDC-E}
        & \underline{91.7(1.1)} & 2.9(1.9) & \underline{97.9(5.3)} & 7.1(1.8) & \underline{112.3(1.7)} & \underline{5.8(1.0)} & \underline{5.6(4.1)}
        & \textbf{87.1(1.5)} & 4.1(1.7) & \underline{95.8(6.6)} & 7.4(1.3) & \underline{78.3(25.7)} & \underline{8.8(2.0)} & 4.9(3.9) \\
        \bottomrule
    \end{tabular}
\end{table*}

As shown in Fig. \ref{fig:compare}, both RUDC-E and RUDC-T exhibit superior learning efficiency. Compared to baselines, RUDC rapidly converge to high episode rewards while maintaining the lowest violation rates, demonstrating that the uncertainty-embedded HOCBF effectively guides safe exploration without hindering learning progress.

Tab. \ref{tab:comparison} summarizes the quantitative results. In the Random Destination task, RUDC-T and RUDC-E achieve the highest SR of 92.3\% and 91.7\%, respectively, outperforming all baselines. Lacking explicit safety considerations, vanilla SAC presents a high VR of 21.1\%, underscoring its vulnerability in interactive environments. Among the safe RL baselines, USL performs best with an 88.5\% SR, but still incurs an 11.5\% VR. Notably, QPSL relies strictly on the linear approximation of safety constraints; this inaccuracy causes its VR to exceed even that of vanilla SAC, despite achieving the highest AES. Conversely, Recovery RL operates on a rigid threshold-triggered recovery mechanism, inducing over-conservatism with a 42.0\% FR. In this task, our proposed framework successfully maintains the lowest VR without sacrificing excessive driving efficiency.

The Dense Unprotected Left Turn task drastically increases traffic density compared to the training distribution, elevating the likelihood of OOD and long-tail events. Under these challenging conditions, the performance of most baseline methods degrades markedly. SAC's VR surges to 35.3\%, while Recovery RL and FAC suffer from elevated FRs of 47.3\% and 15.6\%, respectively, failing to navigate the intersection safely and efficiently. In contrast, our framework synergizes CVaR to optimize worst-case returns with HOCBFs to provide formal safety boundaries. This dual mechanism ensures remarkable robustness for RUDC-E and RUDC-T, sustaining SRs of 87.1\% and 86.3\%.

\subsection{Ablation Studies}

\begin{figure*}
    \centering
    \subfloat{%
        \includegraphics[width=0.6\textwidth]{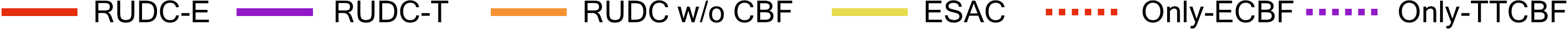}
    }\\ \vspace{-5pt}
    \subfloat{%
        \includegraphics[width=0.24\textwidth]{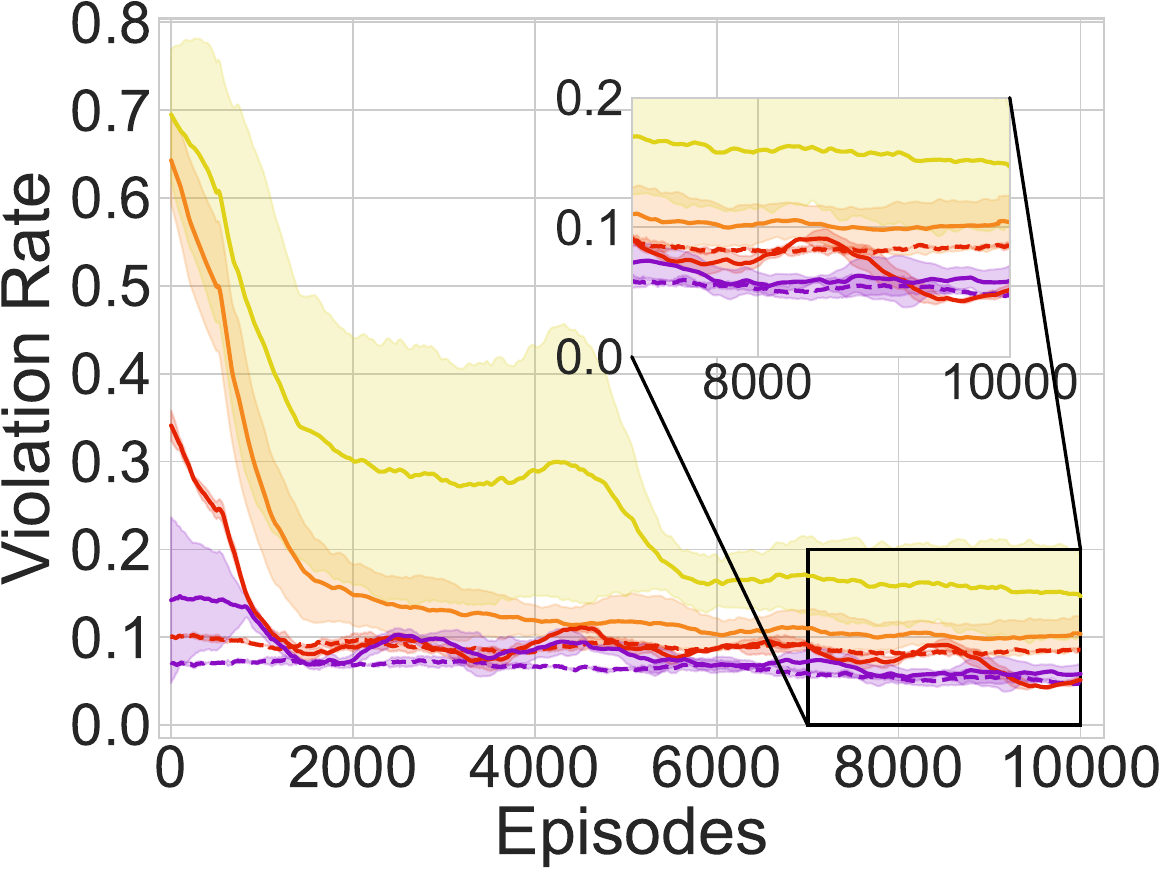}
        \label{ablution_violate}
    }\subfloat{%
        \includegraphics[width=0.24\textwidth]{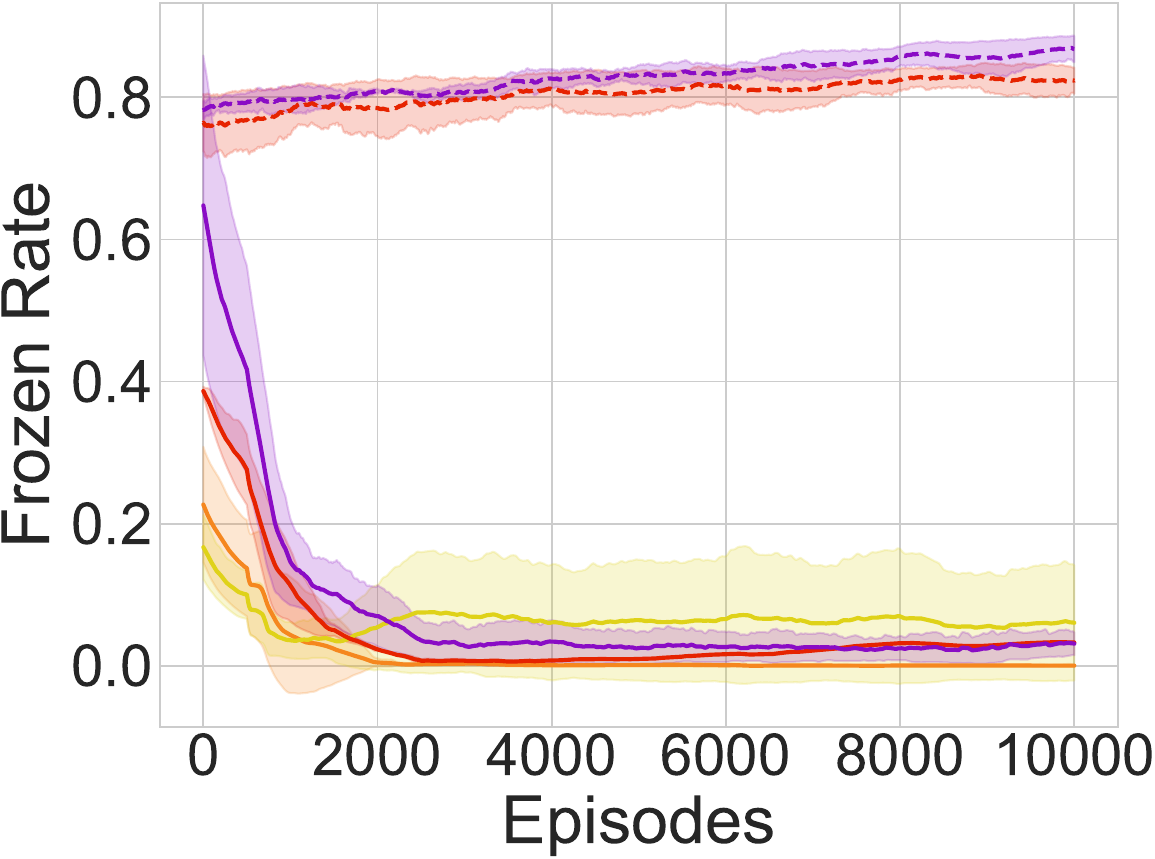}
        \label{ablution_frozen}
    }\subfloat{%
        \includegraphics[width=0.24\textwidth]{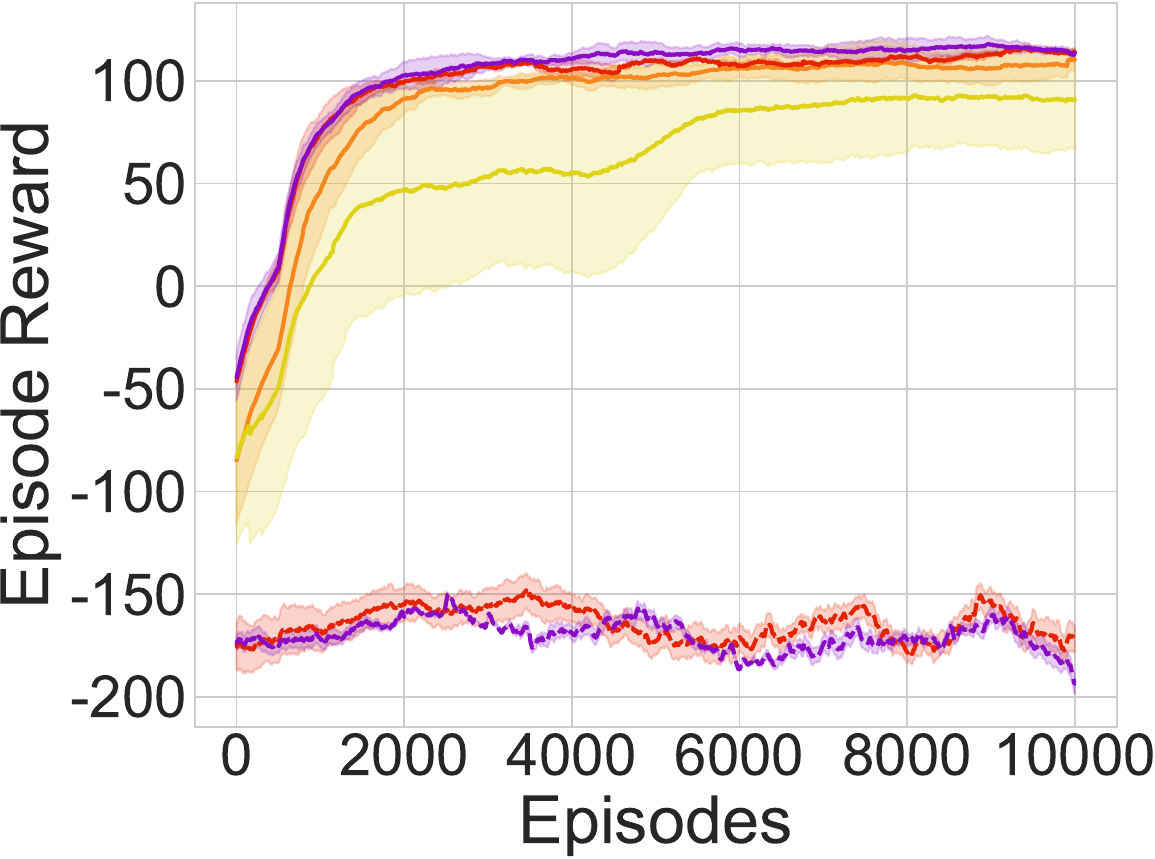}
        \label{ablution_reward}
    }\subfloat{%
        \includegraphics[width=0.24\textwidth]{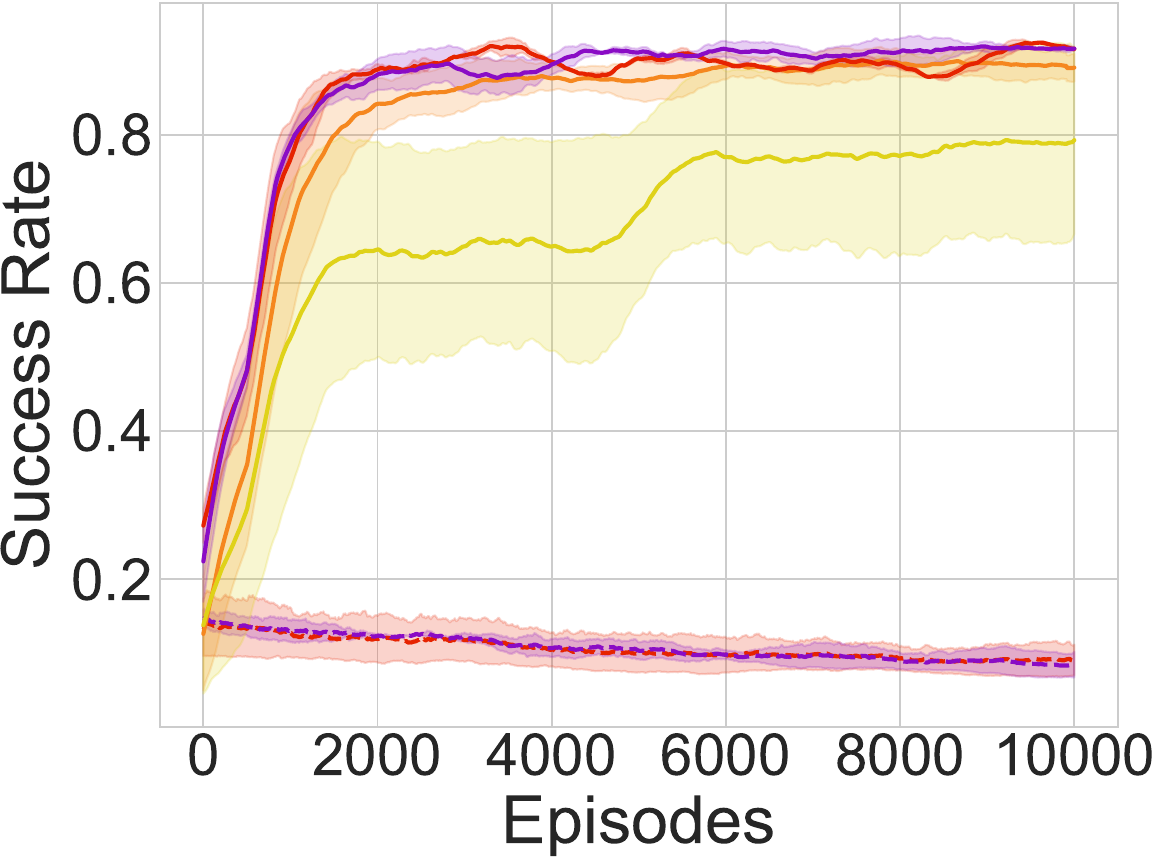}
        \label{ablution_success}
    }
    \caption{Training curves of Ablation studies. Solid lines denote means, while shaded regions represent 95\% confidence intervals over three runs.}
    \label{fig:ablution}
\end{figure*}

\begin{table*}[t]
    \centering
    \scriptsize
    \caption{Ablation study across two Test Conditions. 
    $\uparrow$\,/\,$\downarrow$ indicates higher\,/\,lower is better.
    \textbf{Bold} and \underline{underlined} denote the best and second-best results, respectively. Values are presented as mean (standard deviation).}
    \label{tab:ablation}
    \setlength{\tabcolsep}{2.2pt}
    \renewcommand{\arraystretch}{1.25}
    \begin{tabular}{l c c c c c >{\columncolor{gray!15}}c >{\columncolor{gray!15}}c c c c c c >{\columncolor{gray!15}}c >{\columncolor{gray!15}}c}
        \toprule
        \textbf{Variant} 
        & \multicolumn{7}{c}{\textit{Random Destination}}
        & \multicolumn{7}{c}{\textit{Dense Unprotected Left Turn}} \\
        \cmidrule(lr){2-8} \cmidrule(lr){9-15}
        & SR\,$\uparrow$ & FR\,$\downarrow$ & RC\,$\uparrow$ & AES\,$\uparrow$ & AER\,$\uparrow$ & VR\,$\downarrow$ & min-DTC\,$\uparrow$
        & SR\,$\uparrow$ & FR\,$\downarrow$ & RC\,$\uparrow$ & AES\,$\uparrow$ & AER\,$\uparrow$ & VR\,$\downarrow$ & min-DTC\,$\uparrow$ \\
        & (\%) & (\%) & (\%) & (m/s) & & (\%) & (m)
        & (\%) & (\%) & (\%) & (m/s) & & (\%) & (m) \\
        \midrule
        ESAC 
        & 78.9(14.0) & 5.9(8.4) & 97.1(4.5) & \textbf{7.9(1.5)} & 90.9(26.9) & 15.2(5.7) & 3.1(2.9)
        & 73.2(16.1) & 4.5(7.2) & 88.5(15.2) & \textbf{8.1(1.2)} & 20.5(38.4) & 28.5(9.8) & 1.8(1.9) \\
        
        RUDC w/o CBF 
        & 89.4(2.2) & \textbf{0.1(0.2)} & 95.3(5.2) & \underline{7.8(1.1)} & 106.9(10.6) & 10.1(2.0) & 3.9(2.2)
        & 83.3(8.5) & \textbf{1.2(0.8)} & 93.2(7.1) & \underline{7.5(0.9)} & 45.2(12.5) & 15.5(4.6) & 3.3(1.9) \\
        
        Only-TTCBF 
        & 9.2(2.4) & 86.2(2.2) & 78.5(15.2) & 3.8(0.8) & -175.3(10.5) & \underline{4.6(0.2)} & \textbf{12.5(7.4)}
        & 8.5(3.5) & 88.1(4.5) & 80.1(19.5) & 3.2(0.5) & -210.4(31.4) & \textbf{3.4(0.3)} & \textbf{12.1(8.3)} \\
        
        Only-ECBF 
        & 8.9(3.8) & 82.4(5.8) & 81.8(11.9) & 3.9(0.9) & -163.4(17.2) & 8.7(2.8) & \underline{10.3(7.3)}
        & 9.1(4.2) & 83.5(6.1) & 83.3(10.4) & 3.5(0.6) & -155.1(24.2) & \underline{7.6(3.4)} & \underline{10.9(7.2)} \\
        
        \midrule
        \textbf{RUDC-T}
        & \textbf{92.3(1.2)} & 3.2(1.1) & \textbf{98.5(5.9)} & 7.3(1.3) & \textbf{115.0(4.6)} & \textbf{4.5(0.2)} & 6.3(4.3)
        & \underline{86.3(2.1)} & 5.8(1.2) & \textbf{97.6(6.1)} & 6.7(2.2) & \textbf{83.0(16.2)} & 7.9(1.6) & 5.7(4.9) \\
        \textbf{RUDC-E}
        & \underline{91.7(1.1)} & \underline{2.9(1.9)} & \underline{97.9(5.3)} & 7.1(1.8) & \underline{112.3(1.7)} & 5.8(1.0) & 5.6(4.1)
        & \textbf{87.1(1.5)} & \underline{4.1(1.7)} & \underline{95.8(6.6)} & 7.4(1.3) & \underline{78.3(25.7)} & 8.8(2.0) & 4.9(3.9) \\
        \bottomrule
    \end{tabular}
\end{table*}

\begin{figure*}
    \centering
    \subfloat{%
        \includegraphics[width=0.98\textwidth]{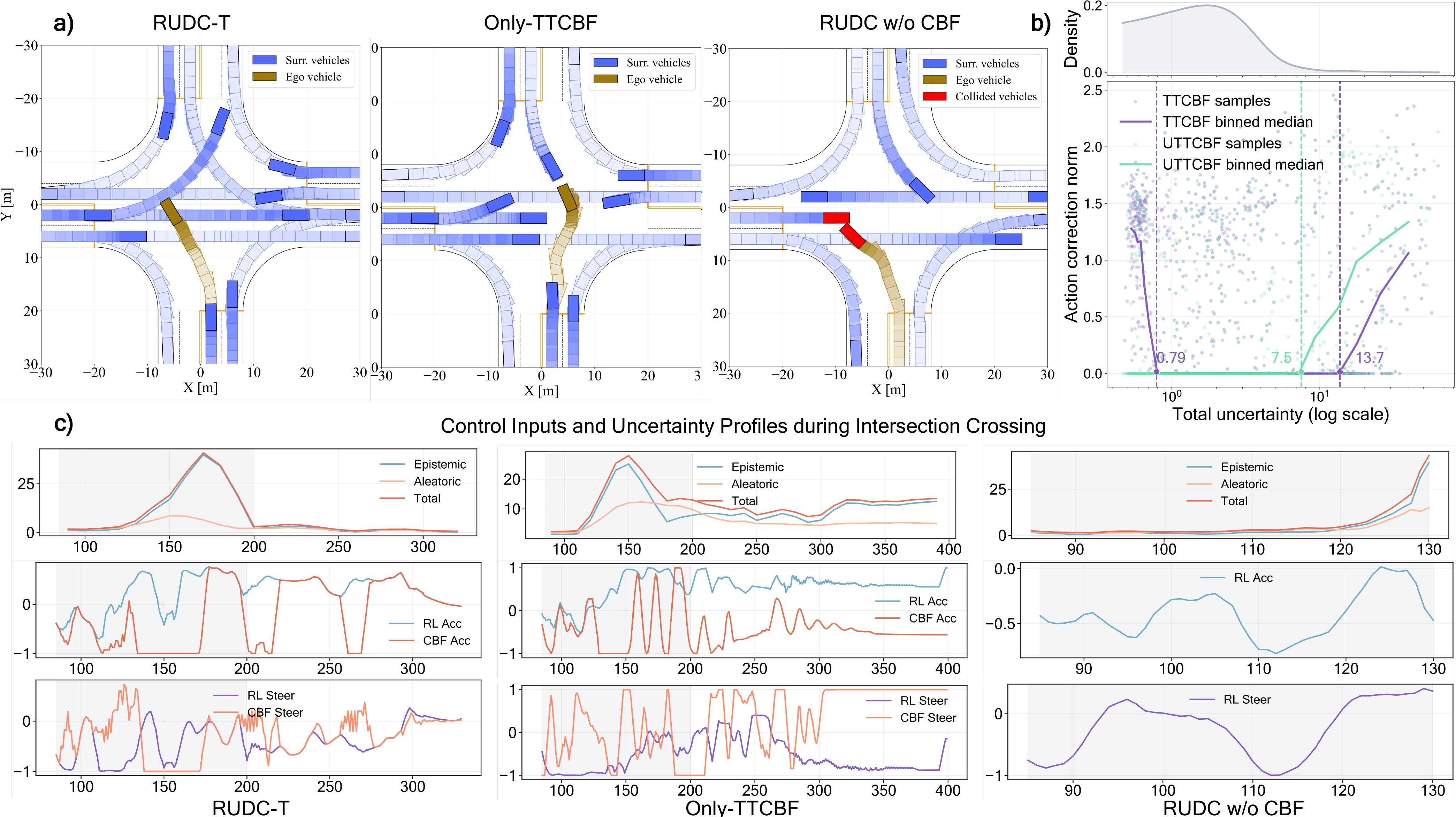}
    }
    \caption{Ablation Case Study on Dense Unprotected Left-Turn Scenarios. (a) Vehicle trajectory visualization, where blue, brown, and red rectangles denote SVs, EVs, and collided vehicles, respectively. (b) 2D distribution of total uncertainty versus CBF-induced action correction magnitude. Numerical annotations indicate the turning points where the median curve significantly exceeds zero. (c) Control inputs and uncertainty profiles during intersection crossing, where the x-axis denotes time steps (0.1s/step) and the y-axes represent uncertainty values, normalized acceleration, and steering angle. The gray shaded area corresponds to the time window visualized in (a).}
    \label{fig:ablution_case}
\end{figure*}

The components of RUDC are validated through a comparative analysis with four ablated variants, as detailed in Tab. \ref{tab:ablation} and Fig. \ref{fig:ablution}. Specifically, Ensemble SAC (ESAC) and RUDC w/o CBF are designated to represent policies lacking HOCBF safety filters, while Only-TTCBF and Only-ECBF served as baselines to isolate the impact of the uncertainty-based constraint adaptation mechanism.
Experimental results reveal that policies without explicit safety filters tend to be overly aggressive; although they achieve higher average speeds, they suffer from severe safety degradation, with VR peaking at 15.5\%. Conversely, while the Only-TTCBF and Only-ECBF variants prioritize safety, the absence of uncertainty-aware adaptation traps the agent in a "conservative deadlock." By enforcing rigid safety margins, these baselines exhibit FR exceeding 78\% and SR below 13\%. Although these pure CBF methods show lower violation rates during the initial training phase, this is a byproduct of operational stagnation; the vehicle avoids collisions by remaining nearly stationary, which ultimately prevents the collection of diverse interactive data for policy improvement. 
In contrast, the full RUDC framework employs uncertainty-embedded boundaries to encourage adaptive exploration. It is noteworthy that in the early stages of training, RUDC variants exhibit higher violation rates than the pure CBF baselines. This phenomenon stems from the adaptive exploration granted by our uncertainty-embedded mechanism, which allows the agent to conduct trial-and-error at the boundaries of the safe set rather than succumbing to the conservative deadlock. As training progresses, this strategic trade-off enables the agent to learn a more robust policy that maintains an SR above 86\% while suppressing the VR to under 10\% in dense scenarios, demonstrating a superior balance between safety-critical requirements and operational flexibility.

We further analyze a case study in the Dense Unprotected Left-Turn task using RUDC-T and its ablation variants (Fig. \ref{fig:ablution_case}). When entering dense traffic, total uncertainty surges across all models, triggering distinct behaviors.  Although RUDC w/o CBF attempts to execute braking commands to prevent an accident, the absence of formal safety constraints renders it incapable of maintaining a reasonable safe distance under high RL policy uncertainty, ultimately resulting in a collision. Conversely, Only-TTCBF exhibits overly conservative behavior. Hampered by the inherent conservatism of the standard HOCBF, the EV fails to accurately identify feasible passing windows. Consequently, it executes an evasive right-turn maneuver, leading to a prolonged standoff with SVs and failing to cross the intersection. In contrast, RUDC-T relaxes the constraints when the RL policy confidence is relatively high, thereby preserving the flexibility of RL decisions. This mechanism empowers the EV to accurately seize passing opportunities and navigate the intersection safely and efficiently. To systematically explain these behaviors, we analyze the distribution of total uncertainty versus the CBF-induced action correction magnitude, as shown in Fig. \ref{fig:ablution_case}(b). The pure TTCBF exhibits an early turning point at a notably low uncertainty level of 0.79, indicating that it aggressively corrects even highly confident RL actions, which inevitably induces over-conservatism.  In contrast, the uncertainty-embedded TTCBF (UTTCBF) shifts the turning point to a lower uncertainty level (7.5) compared to pure TTCBF (13.7). This implies that UTTCBF avoids unnecessary interference during normal conditions, yet remains highly sensitive to severe uncertainties, deploying stronger and more decisive interventions exactly when high safety risks emerge.

\subsection{Case Studies in OOD Scenarios}

\begin{figure*}
    \centering
    \subfloat{%
        \includegraphics[width=0.95\textwidth]{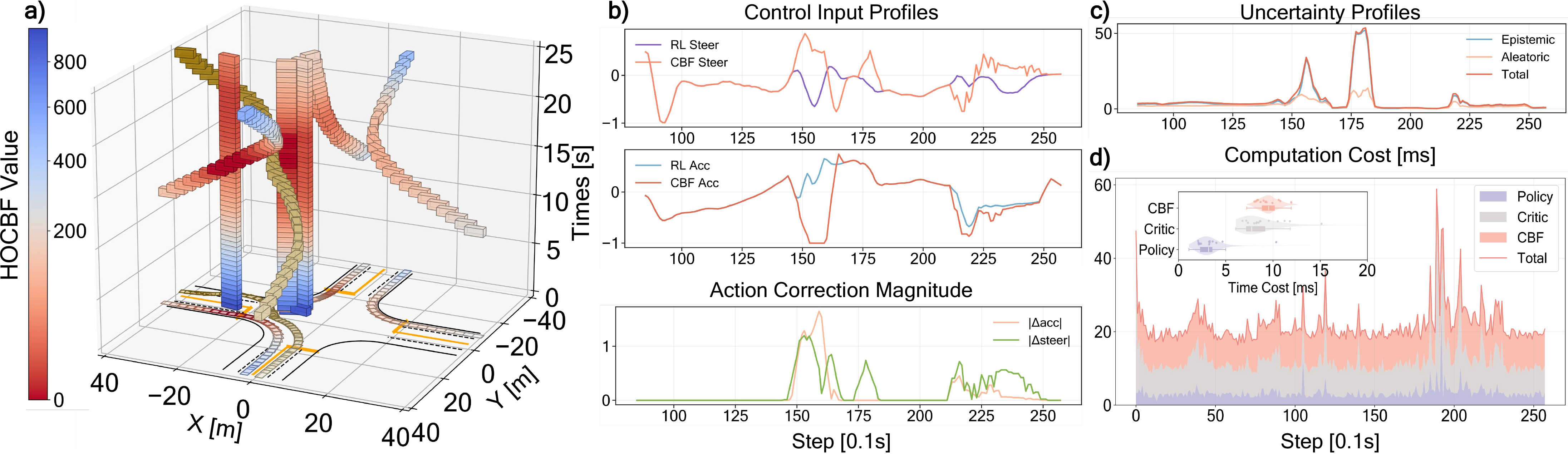}
        \label{ood_case}
    }
    \caption{Case study of an OOD scenario. (a) Spatiotemporal trajectories of the EV and SVs. In this scenario, three stationary vehicles involved in a traffic accident block the intersection, obstructing the EV's normal passage. The color gradient indicates the HOCBF value. (b) Control input profiles and action correction magnitude. (c) Uncertainty profiles. (d) Computational time cost of the CBF, Critic networks, and Policy network.}
    \label{fig:ood_case}
\end{figure*}

In real-world urban driving, intersections are inherently accident-prone. AVs must therefore maintain safe and robust decision-making capabilities even when confronted with unexpected obstacles, such as accident vehicles. Fig. \ref{fig:ood_case} illustrates an unprotected left-turn scenario where three accident vehicles obstruct the nominal path. Given their rarity in both real-world traffic and training datasets, such scenarios constitute a formidable OOD challenge for RL-based decision-making.

The EV approaches the accident zone around step 145. Simultaneously, an oncoming right-turning vehicle encroaches on the EV's vicinity, severely compressing the navigable space. Correspondingly, the $\sigma_{\text{total}}$ experiences a sharp surge, indicating degraded RL reliability. Consequently, a rapid decline in the HOCBF value triggers an increased safety correction on the RL policy. This intervention forces the EV to brake, cautiously creep around the blockade, and yield to the dynamic vehicle. Once the EV successfully bypasses this bottleneck, $\sigma_{\text{total}}$ plummets, prompting the HOCBF to relax its control correction and allowing the EV to resume acceleration.  Subsequently, a third accident vehicle in the exiting lane and a newly merging right-turning vehicle create another spatial bottleneck, causing a second spike in $\sigma_{\text{total}}$. The EV responds by steering left to expand its safety margin, navigating slowly to avoid obstacles and road boundaries, and ultimately reaching its destination safely.

Furthermore, we evaluate the computational overhead during navigation. The experiments are conducted on an AMD EPYC 7542 CPU and an NVIDIA RTX 4090 GPU. The latency primarily originates from the HOCBF optimization and the Critic/Policy network inferences. With a peak cost below 60 ms, the algorithm strictly meets the real-time constraints of autonomous driving (i.e., $\geq 10 Hz$). With an average total latency of only 21.88 ms, the overall computational load is primarily dominated by the HOCBF optimization (approx. 43.8\%) and Critic network inference (approx. 40.7\%), which require comparable execution times. Meanwhile, the Policy network inference contributes minimally, accounting for merely 15.5\% of the total overhead.

\subsection{Sensitivity Analysis}
We conduct sensitivity analyses on two critical hyper-parameters: the $\text{CVaR}$ risk parameter $\beta$, which governs the trade-off between safety and task progress; and the critic ensemble size $N_{\text{ens.}}$, which balances uncertainty estimation quality with computational efficiency.

\subsubsection{Sensitivity to Risk Parameters}
\begin{figure*}
    \centering
    \subfloat{%
        \includegraphics[width=0.95\textwidth]{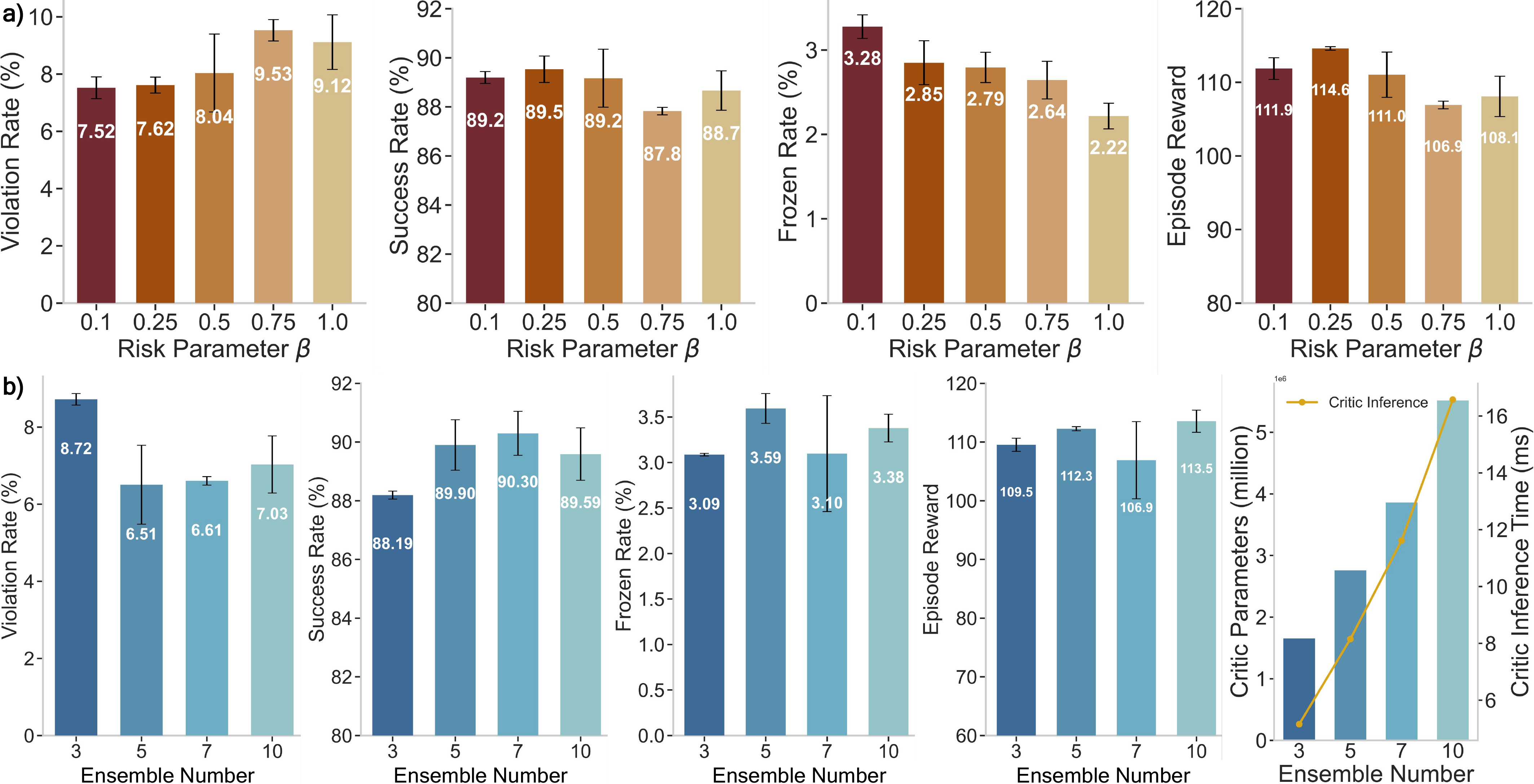}
        \label{sensitivity}
    }
    \caption{Sensitivity analysis with respect to (a) the risk parameter $\beta$ and (b) the critic ensemble size $N_{\text{ens.}}$.}
    \label{fig:sensitivity}
\end{figure*}

We vary the $\text{CVaR}$ risk parameter $\beta$ from 0.1 (highly risk-averse) to 1.0 (risk-neutral). As shown in Fig.~\ref{fig:sensitivity}(a), while a lower $\beta$ generally reduces violations by focusing on worst-case outcomes, extreme risk aversion degrades navigation efficiency. Principally, an excessively small $\beta$  induces over-pessimism: the agent perceives normal states as highly hazardous and freezes to avoid potential penalties. Furthermore, it exacerbates fluctuations in uncertainty estimation. Conversely, a risk-neutral setting $\beta=1.0$  ignores tail risks, leading to a higher violation rate. Consequently, $\beta=0.25$ serves as the best compromise, mitigating tail risks without sacrificing the agent’s ability to complete the task.

\subsubsection{Sensitivity to Ensemble Size}

We analyze the effect of the ensemble size by varying $N_{\text{ens.}}$ from 3 to 10. The results graphed in Fig.~\ref{fig:sensitivity}(b) show that increasing $N_{\text{ens.}}$ improves performance only up to a certain point. Specifically, when $N_{\text{ens.}}$ increases from 3 to 5, both the success rate and the episode reward improve noticeably, while the violation rate is also reduced. However, further increasing $N_{\text{ens.}}$ to 7 or 10 does not bring consistent gains; instead, the improvements become marginal and some metrics even fluctuate or degrade. Meanwhile, the parameter size and inference time grow almost linearly with $N_{\text{ens.}}$, leading to a substantially higher computational cost. Based on these observations, we choose $N_{\text{ens.}}=5$ as the default setting in all experiments. This choice achieves the best overall trade-off among task performance, safety, and efficiency: it provides the strongest empirical performance while avoiding the unnecessary computational overhead introduced by larger ensembles.

\section{Conclusion}
This paper presents the Risk-sensitive and Uncertainty-aware Decision-making and Control (RUDC) framework for safe and robust autonomous driving at highly interactive unsignalized intersections. By intrinsically coupling risk-sensitive distributional reinforcement learning with ensemble-based policy uncertainty quantification, the proposed framework enables reliability-aware decision-making in safety-critical intersection navigation. Furthermore, an uncertainty-aware HOCBF-based safety correction mechanism is introduced to dynamically adjust safety constraints according to policy uncertainty while compensating CBF model mismatches through a learnable residual predictor. Extensive simulations demonstrate that RUDC achieves a superior balance between safety, efficiency, and robustness compared with existing safe RL baselines while satisfying real-time requirements.

Despite these advancements, several avenues remain for future research. First, the current CBF formulation relies exclusively on instantaneous state-action pairs without considering historical context, which can occasionally lead to corrected actions that deviate significantly from the nominal policy. Future work will explore Predictive CBFs to integrate historical information, ensuring smoother and more consistent safety interventions. Second, the integration of input-constrained CBFs will be investigated to address the potential conflicts between safety requirements and physical actuation limits. This is crucial for preventing scenarios where the absence of feasible control inputs within the prescribed bounds violates the forward invariance of the safe set, thereby ensuring rigorous safety guarantees even under strict hardware constraints. Finally, the framework’s robustness and adaptability will be further validated through extensive testing in mixed traffic environments involving heterogeneous traffic participants.

\appendices
\setcounter{table}{0}
\setcounter{figure}{0}
\setcounter{section}{0}
\setcounter{equation}{0}
\setcounter{definition}{0}
\renewcommand{\thetable}{A.\arabic{table}}
\renewcommand{\thefigure}{A.\arabic{figure}}
\renewcommand{\thesection}{A.\arabic{section}}
\renewcommand{\theequation}{A.\arabic{equation}}
\renewcommand{\thedefinition}{A.\arabic{definition}}
\section{Details of High-Order Control Barrier Functions}
\label{apx:hocbf}
As established in the kinematic model, the control input $\boldsymbol{u}$ appears in the second derivative of the candidate barrier function $h_i(\boldsymbol{x}_i)$, indicating a relative degree of $r = 2$. In this appendix, we detail two specific implementations: Exponential Control Barrier Functions (ECBF) \cite{7524935} and Truncated Taylor Control Barrier Functions (TTCBF) \cite{xu2025high}.

\subsection{Exponential Control Barrier Functions}
The ECBF approach enforces high relative-degree safety constraints by utilizing concepts from linear control theory and therefore conventional methods such as pole placement control can be used to design ECBF constraints.

\begin{definition} [ECBF \cite{7524935}]
Define the state vector of the barrier function as $\boldsymbol{\eta}_b(\boldsymbol{x}) = [h(\boldsymbol{x}), \dot{h}(\boldsymbol{x}), \dots, h^{(r-1)}(\boldsymbol{x})]^\top$. Given a set $\mathcal{C}$ as in (\ref{eq:safe_set_define}), a continuously differentiable function $h$ is a candidate ECBF with relative degree $r$ if there exist $\boldsymbol{K}_b \in \mathbb{R}^{1 \times r}$ such that $\forall \boldsymbol{x} \in \mathcal{C}$,

\begin{equation}
\begin{aligned}[b]
\begin{gathered}
\sup_{\boldsymbol{u} \in \mathcal{U}} [L_f^r h(\boldsymbol{x}) + L_g L_f^{r-1} h(\boldsymbol{x}) \boldsymbol{u} + \boldsymbol{K}_b \boldsymbol{\eta}_b(\boldsymbol{x})] \ge 0
\end{gathered}
\end{aligned}
\end{equation} where $\boldsymbol{K}_b$ is chosen such that the eigenvalues of the matrix $(A_b - B_b \boldsymbol{K}_b)$ are real and negative, ensuring $h(\boldsymbol{x}(t)) \ge \boldsymbol{C}_b e^{(A_b - B_b \boldsymbol{K}_b)t} \boldsymbol{\eta}_b(\boldsymbol{x}_0) \ge 0$, where $A_b$ is the nilpotent shift matrix with ones on the superdiagonal, and $B_b = e_n = [0,\dots,0,1]^\top$.
\end{definition} 

For $r = 2$, we define the gains based on poles $-\lambda_1^E, -\lambda_2^E$ (where $\lambda_1^E, \lambda_2^E > 0$). The constraint can be reformulated as:
\begin{equation}
\begin{aligned}[b]
\sup_{\boldsymbol{u}_k\in\mathcal{U}}[\ddot{h}_i(\boldsymbol{x}_i, \boldsymbol{u}) + (\lambda_1^E + \lambda_2^E)\dot{h}_i(\boldsymbol{x}_i) + \lambda_1^E \lambda_2^E h_i(\boldsymbol{x}_i)] \ge 0.
\end{aligned}
\end{equation}

ECBF requires $\lambda_1^E \geq -\dot h(x_0)/h(x_0)$ to ensure initial feasibility \cite{xu2025high}. However, satisfying this state-dependent condition continuously is computationally intractable in highly dynamic and complex interactive scenarios. Following \cite{xu2025high}, we empirically tuned these parameters via a systematic grid search in simulations. By evaluating parameter combinations within the space $(\lambda_1^E,\lambda_2^E)\in[0.1, 3.0]\times[0.1, 3.0]$, we selected $\lambda_1^E=1.5, \lambda_2^E=0.3$ as the final configuration, as it yielded the minimum violation rate.

\subsection{Truncated Taylor Control Barrier Functions}
To mitigate the tuning complexity associated with multiple parameters in the ECBF approach, TTCBF approximates the discrete-time CBF condition using a truncated Taylor series and requires only a class $\mathcal{K}$ function. For a system with relative degree $r$, TTCBF approximates $\Delta h(\boldsymbol{x}_k,\boldsymbol{u}_k)$ in (\ref{eq:dis_cbf}) as $\Delta h(\boldsymbol{x}_k,\boldsymbol{u}_k)\approx \Delta t\dot{h}(\boldsymbol{x}_k)+\frac{1}{2}\Delta t^2\ddot{h}(\boldsymbol{x}_k)+\cdots+\frac{1}{r!}\Delta t^rh^{(r)}(\boldsymbol{x}_k,\boldsymbol{u}_k)$, with the $r$-th derivative $h^{(r)}(\boldsymbol{x}_k,\boldsymbol{u}_k)$ capture the control input.
\begin{definition} [TTCBF \cite{xu2025high}]
Given a set $\mathcal{C}$ as in (\ref{eq:safe_set_define}), a continuously differentiable function $h$ is a candidate TTCBF with relative degree $r$ if there exist class $\mathcal{K}$ functions $\alpha(x)\leq x$ such that $\forall \boldsymbol{x} \in \mathcal{C}$,
\begin{equation}
\begin{aligned}[b]
\begin{gathered}
\sup_{\boldsymbol{u}_k\in\mathcal{U}}[\Delta t\dot{h}(\boldsymbol{x}_k)+\cdots+\frac{1}{r!}\Delta t^rh^{(r)}(\boldsymbol{x}_k,\boldsymbol{u}_k)+ \\\alpha\left(h(\boldsymbol{x}_k)\right)]\geq\Gamma\Delta t^{r+1},
\end{gathered}
\end{aligned}
\end{equation} where $\Gamma$ is a parameter satisfying $\frac{\Gamma} {(r+1)!}t^{r+1}\geq|R_{r+1}|=\frac{|h^{(r+1)}(\boldsymbol{\xi})|} {(r+1)!}t^{r+1}, \boldsymbol{\xi}\in[\boldsymbol{x}_{k},\boldsymbol{x}_{k+1}]$.
\end{definition}

For $r = 2$, applying a linear class $\mathcal{K}$ function $\alpha(h) = \lambda_1^T h$ with a single parameter $\lambda_1^T \in (0, 1]$, the TTCBF constraint is constructed as:
\begin{equation}
\begin{aligned}[b]
\sup_{\boldsymbol{u}_k\in\mathcal{U}}[\Delta t \dot{h}(\boldsymbol{x}_{k}) + \frac{1}{2}\Delta t^2 \ddot{h}(\boldsymbol{x}_{k}, \boldsymbol{u}_k) + \lambda_1^T h(\boldsymbol{x}_{k})] \ge \Gamma \Delta t^3
\end{aligned}
\end{equation} where $\Delta t$ represents the discrete sampling period. Similar to the parameter selection of ECBF, we evaluate $\lambda_1^T$ within the range of [0.05, 1.0] and finally set it to 0.1.

\section{Validation of CBF residual predictor}
\label{apx:cbf_predictor}

\begin{figure}
    \centering
    \subfloat{%
        \includegraphics[width=0.3\textwidth]{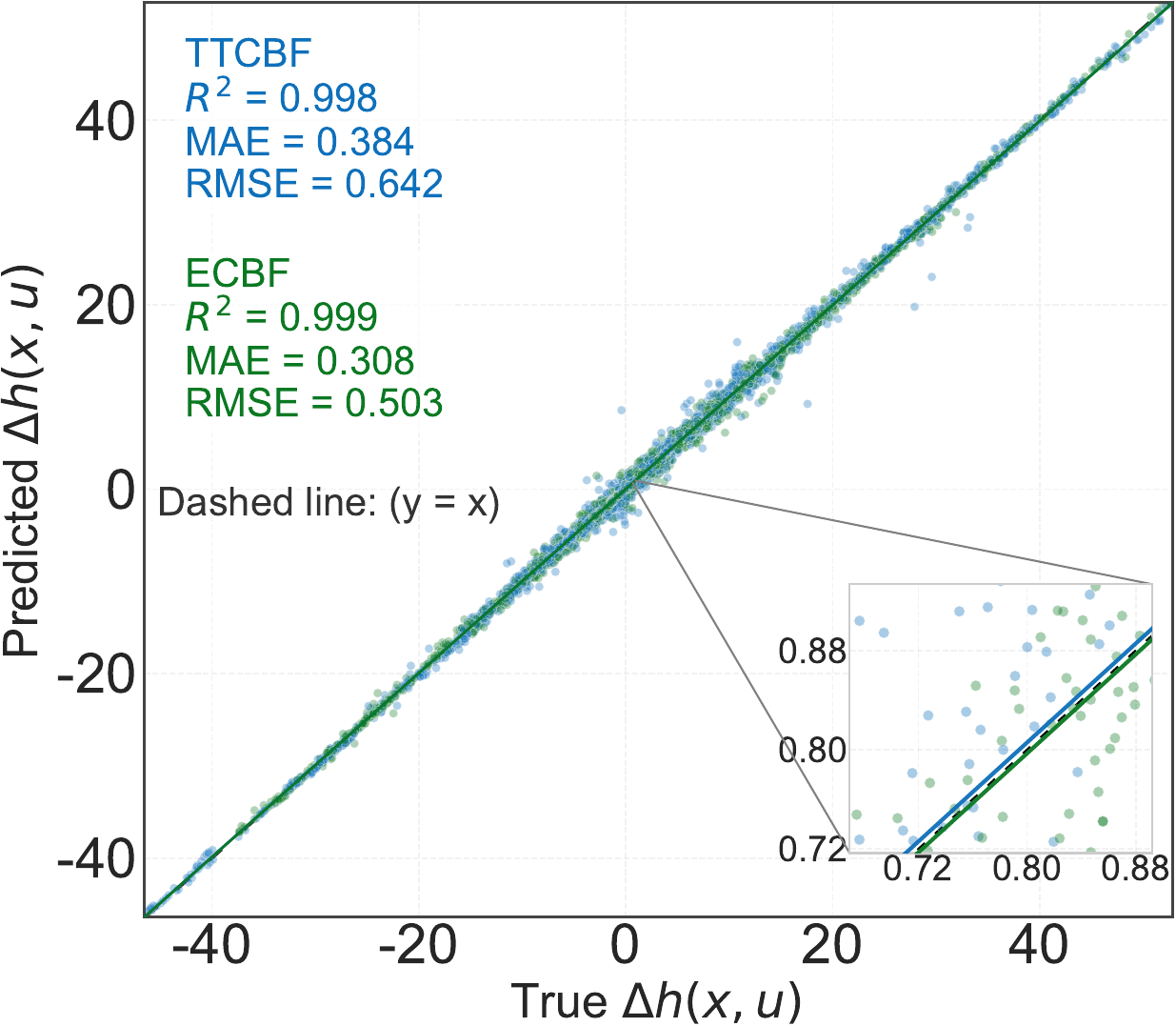}
        \label{ttcbf_ecbf_delta_h}
    }
    \caption{Regression performance of the MLP-based residual predictors for TTCBF and ECBF. The scatter plots compare the network's predicted barrier variations against the true observed values. The tight clustering along the y=x (dashed) line and high $R^2$ scores ($\ge0.98$) demonstrate the predictor's exceptional accuracy in capturing model mismatches.}
    \label{fig:cbf_predictor}
\end{figure}

As introduced in Sec. \ref{sec:ue_cons}, accurately estimating the safety boundary residual $w_k$ is crucial for compensating model mismatches and discretization errors in HOCBFs. We evaluate the proposed MLP predictor on an offline dataset of $10^5$ random exploration samples. The regression performance is quantitatively assessed using three standard metrics: R-squared ($R^2$), Mean Absolute Error (MAE), and Root Mean Squared Error (RMSE). As shown in Fig. \ref{fig:cbf_predictor}, the predicted $\Delta h(\boldsymbol{x}, \boldsymbol{u})$ closely align with the true values along the $y=x$ reference line. Quantitatively, the predictor achieves high accuracy for both TTCBF and ECBF. This demonstrates the predictors' capability to capture unmodeled dynamics and errors, enabling the uncertainty-embedded HOCBF to dynamically modulate safety constraints online without suffering from over-conservatism or safety violations.

\section{Safe Reinforcement Learning}
\label{apx:saferl}
Safe RL is usually modeled as Constrained Markov Decision Process (CMDP). Extending the standard MDP, a CMDP incorporates safety constraints and is formally defined by a tuple $(\mathcal{S}, \mathcal{A}, \mathcal{P}, r, \rho_0, \gamma, \mathcal{C})$, where $\mathcal{C}:\mathcal{S}\times{\mathcal{A}}\rightarrow[0,+\infty]$ maps the state action transition tuple into a cost value and reflects the constraint violation. In contrast to a standard MDP, a CMDP requires optimizing the reward while adhering to safety constraints. Consequently, CMDP can be formulated as the following constrained optimization problem:
\begin{equation}
\begin{aligned}[b]
\mathcal{L}_{R}(\pi)=\mathbb{E}_{\pi,\mathcal{P}}\!\left[\sum_{t=0}^{\infty}\gamma^tr(s_t,a_t)\right],
s.t.\mathcal{L}_{C}(\pi)\leq d_{\text{th}}, 
\end{aligned}
\label{eq:constrained_rl}
\end{equation} where $\mathcal{L}_{C}(\pi)=\mathbb{E}_{\pi,\mathcal{P}}\left[\sum_{t=0}^\infty\gamma^tc(s_t,a_t)\right]$ quantifies the expected cumulative cost incurred by policy $\pi$ under safety constraints, $d_{\text{th}} \in \mathbb R$ is the constraint threshold. We briefly introduce the core mathematical mechanisms of the selected baselines below.
\vspace{-5pt}
\subsection{Lagrangian Relaxation}
This approach reformulates the constrained RL problem defined in \ref{eq:constrained_rl} via Lagrangian relaxation, transforming the original CMDP into a primal-dual saddle-point problem:
\begin{equation}
\max_{\lambda \geq 0} \min_{\theta} \mathbb{E}_{\mathcal{D}} \left[ - Q^\pi(s, \pi_\theta(s)) + \lambda \left( Q_c^\pi(s, \pi_\theta(s)) - d_{\text{th}} \right) \right].
\end{equation} where $\lambda$ is updated via dual ascent methods.

\vspace{-5pt}
\subsection{Feasible Actor Critic}
To address the limitation of trajectory-averaged constraints in standard Lagrangian methods, FAC introduces a state-dependent multiplier $\lambda(s)$ to enforce state-wise safety:
\begin{equation}
\max_{\lambda \geq 0} \min_{\theta} \mathbb{E}_{\mathcal{D}} \left[ - Q^\pi(s, \pi_\theta(s)) + \lambda(s) \left( Q_c^\pi(s, \pi_\theta(s)) - d_{\text{th}} \right) \right].
\end{equation}
\vspace{-25pt}
\subsection{Safety Layer}
SL integrates a post-processing module that projects potentially unsafe actions onto a locally safe half-space. It uses a learned linear approximation of the cost, $C(s_t, a_t) \approx g(s_t; \omega)^\top a_t + c_{t-1}$, and solves a Quadratic Program (QP) at each step:
\begin{equation}
a_t^* = \arg\min_a \frac{1}{2} | a - \mu_\theta(s) |^2 \quad \text{s.t.} g(s_t; \omega)^\top a_t + c_{t-1} \leq d_{\text{th}}.
\end{equation}
\vspace{-25pt}
\subsection{Unrolling Safety Layer}
The USL addresses state-wise hard constraints by integrating safety optimization with iterative projection. Specifically, it employs a deep unrolling architecture to refine the policy’s initial output $a_0$ via a gradient-based correction operator:
\begin{equation}
\psi(a^k) = a^k - \frac{\eta}{\mathcal{Z}k} \nabla{a^k} [Q_c(s, a^k) - d_{\text{th}}]^+,
\end{equation}
where $\delta$ is the safety threshold and $\mathcal{Z}_k = \| \nabla_{a^k} [Q_c(s, a^k) - d_{\text{th}}]^+ \|_\infty$ serves as a normalization factor.
\vspace{-15pt}
\subsection{Safety Recovery}
Safety recovery decouples task execution from high-risk interventions. A safety critic estimates the discounted probability of future violations, denoted as $Q_{\text{risk}}^\pi$. A dedicated recovery policy $\pi_{\text{recov}}$ takes over when the estimated risk exceeds a threshold $\delta$:
\begin{equation}
a_t =
\begin{cases}
\pi_{\text{task}}(s_t), & \text{if } Q_{\text{risk}}^{\pi}(s_t, \pi_{\text{task}}(s_t)) \leq d_{\text{th}}, \\
\pi_{\text{recov}}(s_t), & \text{otherwise}.
\end{cases}
\end{equation}
If the task action falls outside this set, a recovery policy is activated to minimize $Q_{\mathrm{risk}}$. 

In practice, we employ a twin-Critic structure for the cost function $Q_c$ (i.e., $Q_{\text{risk}}$). Taking the maximum value $Q_c = \max(Q_{c1}, Q_{c2})$ helps avoid underestimating constraint violations and provides a sufficient safety margin. The detailed hyper-parameters are listed in Tab.~\ref{tab:hyperparameters}.

For the CMDP-based formulation, we construct a unified step-wise cost function that integrates continuous interactive risks and discrete event penalties:
\begin{equation} C = w_d c_{\mathrm{dist}} + w_c \mathbb{I}{\mathrm{col}} + w_c \mathbb{I}{\mathrm{road}} \end{equation}where continuous risk term $c_{\mathrm{dist}}$ is mapped from a normalized metric $z_q = \operatorname{clip}((h_q - d_q)/(h_q - l_q), 0, 1)$ for channels $q \in \{\mathrm{veh}, \mathrm{road}\}$. In our implementation, the safety thresholds are set to $l_q=1.0$ and $h_q=3.0$, with corresponding weights $w_d=1.0$ and $w_c=5.0$.

\begin{table}[h]
    \centering
    \caption{Hyper-parameters of different Safe RL algorithms}
    \label{tab:hyperparameters}
    \scriptsize
    \setlength{\tabcolsep}{4pt}
    \renewcommand{\arraystretch}{1.1}
    \begin{tabular}{lccccc}
        \hline
        Hyper-parameters & QPSL & Recovery RL & Lagrangian & FAC & USL \\
        \midrule
        Cost Limit $d_{\text{th}}$ & 0.2 & 0.2 & 0.2 & 0.2 & 0.2 \\
        Reward Discount & 0.99 & 0.99 & 0.99 & 0.99 & 0.99 \\
        Cost Discount & 0.99 & 0.99 & 0.99 & 0.99 & 0.99 \\
        Batch Size & 256 & 256 & 256 & 256 & 256 \\
        Critic LR & 3E-3 & 3E-3 & 3E-3 & 3E-3 & 3E-3 \\
        Actor LR & 3E-4 & 3E-4 & 3E-4 & 3E-4 & 3E-4 \\
        Safe Critic LR & 3E-3 & 3E-3 & 3E-3 & 3E-3 & 3E-3 \\
        Safe Actor LR & N/A & 3E-4 & N/A & N/A & N/A \\
        Multiplier LR & N/A & N/A & 1E-5 & 1E-6 & N/A \\
        Multiplier Init & N/A & N/A & 0.0 & N/A & N/A \\
        Multiplier Delay & N/A & N/A & N/A & 6 & N/A \\
        Penalty Factor $\kappa$ & N/A & N/A & N/A & N/A & 5 \\
        Iterative Step $K$ & N/A & N/A & N/A & N/A & 20 \\
        \bottomrule
    \end{tabular}
\end{table}

\bibliographystyle{IEEEtran}
\bibliography{refs}

@ARTICLE{LiTITS2026,
  author={Li, Zhuoren and Leng, Bo and Xiong, Lu and Eichberger, Arno and Huang, Chao and Hu, Jia},
  journal={IEEE Transactions on Intelligent Transportation Systems}, 
  title={Safety-Enhanced Deep Reinforcement Learning for Autonomous Driving: Dare to Make Mistakes to Learn Better and Faster}, 
  year={2026},
  volume={},
  number={},
  pages={1-13},
  doi={10.1109/TITS.2026.3670584}}

@article{wang2024sotifsurvey,
  author = {Hao Wang and Wenshuo Shao and Chao Sun and Kan Yang and Dongpu Cao and Jun Li},
  title = {A Survey on an Emerging Safety Challenge for Autonomous Vehicles: Safety of the Intended Functionality},
  journal = {Engineering},
  volume = {33},
  pages = {17--34},
  year = {2024},
  doi = {10.1016/j.eng.2024.02.001}
}

@article{xiao2022hocbf,
  author = {Wei Xiao and Calin Belta},
  title = {High-Order Control Barrier Functions},
  journal = {IEEE Trans. Autom. Control},
  volume = {67},
  number = {7},
  pages = {3655--3662},
  year = {2022},
  doi = {10.1109/TAC.2021.3105491}
}

@article{xiong2023dtcbf,
  author = {Yong Xiong and Dong-Hua Zhai and Mahdi Tavakoli and Yuanqing Xia},
  title = {Discrete-Time Control Barrier Function: High-Order Case and Adaptive Case},
  journal = {IEEE Trans. Cybern.},
  volume = {53},
  number = {5},
  pages = {3231--3239},
  year = {2023},
  doi = {10.1109/TCYB.2022.3170607}
}

@article{hullermeier2021uncertainty,
  author = {Eyke H{\"u}llermeier and Willem Waegeman},
  title = {Aleatoric and Epistemic Uncertainty in Machine Learning: An Introduction to Concepts and Methods},
  journal = {Mach. Learn.},
  volume = {110},
  number = {3},
  pages = {457--506},
  year = {2021},
  doi = {10.1007/s10994-021-05946-3}
}

@article{10073955,
  author = {Hoel, Carl-Johan and Wolff, Krister and Laine, Leo},
  journal = {IEEE Trans. Intell. Transp. Syst.},
  title = {Ensemble Quantile Networks: Uncertainty-Aware Reinforcement Learning With Applications in Autonomous Driving},
  year = {2023},
  volume = {24},
  number = {6},
  pages = {6030-6041},
  doi = {10.1109/TITS.2023.3251376}
}

@article{ma2025dsac,
  author = {Xiaoteng Ma and Li Xia and Zhengyuan Zhou and Jun Yang and Qianchuan Zhao},
  title = {Distributional Soft Actor-Critic for Risk-Sensitive Reinforcement Learning},
  journal = {J. Artif. Intell. Res.},
  volume = {83},
  pages = {1117--1166},
  year = {2025}
}

@article{rockafellar2002cvar,
  author = {R. Tyrrell Rockafellar and Stanislav Uryasev},
  title = {Conditional Value-at-Risk for General Loss Distributions},
  journal = {J. Bank. Financ.},
  volume = {26},
  number = {7},
  pages = {1443--1471},
  year = {2002},
  doi = {10.1016/S0378-4266(02)00271-6}
}

@inproceedings{osband2016bootstrapped,
  author = {Ian Osband and Charles Blundell and Alexander Pritzel and Benjamin Van Roy},
  title = {Deep Exploration via Bootstrapped {DQN}},
  booktitle = {Adv. Neural Inf. Process. Syst.},
  volume = {29},
  year = {2016}
}

@inproceedings{osband2018prior,
  author = {Ian Osband and John Aslanides and Albin Cassirer},
  title = {Randomized Prior Functions for Deep Reinforcement Learning},
  booktitle = {Adv. Neural Inf. Process. Syst.},
  volume = {31},
  year = {2018}
}

@article{10740674,
  author = {Yang, Kai and Li, Shen and Chen, Yongli and Cao, Dongpu and Tang, Xiaolin},
  journal = {IEEE Trans. Veh. Technol.},
  title = {Towards Safe Decision-Making for Autonomous Vehicles at Unsignalized Intersections},
  year = {2025},
  volume = {74},
  number = {3},
  pages = {3830-3842}
}

@article{10155311,
  author = {Tang, Xiaolin and Zhong, Guichuan and Li, Shen and Yang, Kai and Shu, Keqi and Cao, Dongpu and Lin, Xianke},
  journal = {IEEE Trans. Intell. Transp. Syst.},
  title = {Uncertainty-Aware Decision-Making for Autonomous Driving at Uncontrolled Intersections},
  year = {2023},
  volume = {24},
  number = {9},
  pages = {9725-9735},
  doi = {10.1109/TITS.2023.3283019}
}

@article{10104197,
  author = {Zhou, Weitao and Cao, Zhong and Deng, Nanshan and Jiang, Kun and Yang, Diange},
  journal = {IEEE Trans. Intell. Transp. Syst.},
  title = {Identify, Estimate and Bound the Uncertainty of Reinforcement Learning for Autonomous Driving},
  year = {2023},
  volume = {24},
  number = {8},
  pages = {7932-7942},
  doi = {10.1109/TITS.2023.3266885}
}

@article{10534899,
  author = {Zhang, Zheng and Liu, Qi and Li, Yanjie and Lin, Ke and Li, Linyu},
  journal = {IEEE Trans. Intell. Transp. Syst.},
  title = {Safe Reinforcement Learning in Autonomous Driving With Epistemic Uncertainty Estimation},
  year = {2024},
  volume = {25},
  number = {10},
  pages = {13653-13666},
  doi = {10.1109/TITS.2024.3397700}
}

@inproceedings{10422331,
  author = {Li, Zhuoren and Xiong, Lu and Leng, Bo and Xu, Puhang and Fu, Zhiqing},
  booktitle = {Proc. IEEE Intell. Transp. Syst. Conf.},
  title = {Safe Reinforcement Learning of Lane Change Decision Making with Risk-Fused Constraint},
  year = {2023},
  pages = {1313-1319},
  doi = {10.1109/ITSC57777.2023.10422331}
}

@article{20233614676836,
  title = {On-Ramp Merging for Highway Autonomous Driving: An Application of a New Safety Indicator in Deep Reinforcement Learning},
  journal = {Automot. Innov.},
  author = {Li, Guofa and Zhou, Weiyan and Lin, Siyan and Li, Shen and Qu, Xingda},
  volume = {6},
  number = {3},
  year = {2023},
  pages = {453 - 465},
  issn = {20964250}
}

@article{20244717409481,
  title = {Double Deep Q-Networks Based Game-Theoretic Equilibrium Control of Automated Vehicles at Autonomous Intersection},
  journal = {Automot. Innov.},
  author = {Hu, Haiyang and Chu, Duanfeng and Yin, Jianhua and Lu, Liping},
  volume = {7},
  number = {4},
  year = {2024},
  pages = {571 - 587},
  issn = {20964250}
}

@article{10675394,
  author = {Gu, Shangding and Yang, Long and Du, Yali and Chen, Guang and Walter, Florian and Wang, Jun and Knoll, Alois},
  journal = {IEEE Trans. Pattern Anal. Mach. Intell.},
  title = {A Review of Safe Reinforcement Learning: Methods, Theories, and Applications},
  year = {2024},
  volume = {46},
  number = {12},
  pages = {11216-11235},
  doi = {10.1109/TPAMI.2024.3457538}
}

@inproceedings{stooke2020responsive,
  title = {Responsive safety in reinforcement learning by pid lagrangian methods},
  author = {Stooke, Adam and Achiam, Joshua and Abbeel, Pieter},
  booktitle = {Int. Conf. Mach. Learn.},
  pages = {9133--9143},
  year = {2020},
  organization = {PMLR}
}

@inproceedings{honari2024meta,
  title = {Meta SAC-Lag: Towards Deployable Safe Reinforcement Learning via MetaGradient-based Hyperparameter Tuning},
  author = {Honari, Homayoun and Enayati, Amir M Soufi and Tamizi, Mehran Ghafarian and Najjaran, Homayoun},
  booktitle = {Proc. IEEE/RSJ Int. Conf. Intell. Robots Syst.},
  pages = {619--626},
  year = {2024},
  organization = {IEEE}
}

@inproceedings{achiam2017constrained,
  title = {Constrained policy optimization},
  author = {Achiam, Joshua and Held, David and Tamar, Aviv and Abbeel, Pieter},
  booktitle = {Int. Conf. Mach. Learn.},
  pages = {22--31},
  year = {2017},
  organization = {PMLR}
}

@article{zhang2020first,
  title = {First order constrained optimization in policy space},
  author = {Zhang, Yiming and Vuong, Quan and Ross, Keith},
  journal = {Adv. Neural Inf. Process. Syst.},
  volume = {33},
  pages = {15338--15349},
  year = {2020}
}

@article{shixin2024unmanned,
  title = {The unmanned vehicle on-ramp merging model based on AM-MAPPO algorithm},
  author = {Shixin, Zhao and Feng, Pan and Anni, Jiang and Hao, Zhang and Qiuqi, Gao},
  journal = {Sci. Rep.},
  volume = {14},
  number = {1},
  pages = {19416},
  year = {2024},
  publisher = {Nature Publishing Group UK London}
}

@article{dalal2018safe,
  title = {Safe exploration in continuous action spaces},
  author = {Dalal, Gal and Dvijotham, Krishnamurthy and Vecerik, Matej and Hester, Todd and Paduraru, Cosmin and Tassa, Yuval},
  journal = {arXiv preprint arXiv:1801.08757},
  year = {2018}
}

@inproceedings{zhang2023evaluating,
  title = {Evaluating model-free reinforcement learning toward safety-critical tasks},
  author = {Zhang, Linrui and Zhang, Qin and Shen, Li and Yuan, Bo and Wang, Xueqian and Tao, Dacheng},
  booktitle = {Proc. AAAI Conf. Artif. Intell.},
  volume = {37},
  number = {12},
  pages = {15313--15321},
  year = {2023}
}

@article{10402567,
  author = {Ma, Haitong and Liu, Changliu and Li, Shengbo Eben and Zheng, Sifa and Sun, Wenchao and Chen, Jianyu},
  journal = {IEEE Trans. Neural Netw. Learn. Syst.},
  title = {Learn Zero-Constraint-Violation Safe Policy in Model-Free Constrained Reinforcement Learning},
  year = {2025},
  volume = {36},
  number = {2},
  pages = {2327-2341},
  doi = {10.1109/TNNLS.2023.3348422}
}

@inproceedings{lagrangian,
  title = {Learning to Walk in the Real World with Minimal Human Effort},
  author = {Ha, Sehoon and Xu, Peng and Tan, Zhenyu and Levine, Sergey and Tan, Jie},
  booktitle = {Proc. Conf. Robot Learn.},
  pages = {1110--1120},
  year = {2021},
  volume = {155},
  month = {16--18 Nov},
  publisher = {PMLR},
  url = {https://proceedings.mlr.press/v155/ha21c.html}
}

@inproceedings{10610959,
  author = {Zheng, Haotian and Ma, Haitong and Zheng, Sifa and Li, Shengbo Eben and Wang, Jianqiang},
  booktitle = {IEEE Int. Conf. Robot. Autom.},
  title = {Synthesize Efficient Safety Certificates for Learning-Based Safe Control using Magnitude Regularization},
  year = {2024},
  pages = {545-551},
  doi = {10.1109/ICRA57147.2024.10610959}
}

@article{9718195,
  author = {Wang, Xiao},
  journal = {IEEE Robot. Autom. Lett.},
  title = {Ensuring Safety of Learning-Based Motion Planners Using Control Barrier Functions},
  year = {2022},
  volume = {7},
  number = {2},
  pages = {4773-4780},
  doi = {10.1109/LRA.2022.3152313}
}

@ARTICLE{caozhongTISconfidence,
  author={Cao, Zhong and Xu, Shaobing and Peng, Huei and Yang, Diange and Zidek, Robert},
  journal={IEEE Trans. Intell. Transp. Syst.}, 
  title={Confidence-Aware Reinforcement Learning for Self-Driving Cars}, 
  year={2022},
  volume={23},
  number={7},
  pages={7419-7430}}

@article{10107652,
  author = {Yang, Kai and Tang, Xiaolin and Qiu, Sen and Jin, Shufeng and Wei, Zichun and Wang, Hong},
  journal = {IEEE Trans. Veh. Technol.},
  title = {Towards Robust Decision-Making for Autonomous Driving on Highway},
  year = {2023},
  volume = {72},
  number = {9},
  pages = {11251-11263},
  doi = {10.1109/TVT.2023.3268500}
}

@inproceedings{dabney2018distributional,
  title = {Distributional reinforcement learning with quantile regression},
  author = {Dabney, Will and Rowland, Mark and Bellemare, Marc and Munos, R{\'e}mi},
  booktitle = {Proc. AAAI Conf. Artif. Intell.},
  volume = {32},
  number = {1},
  year = {2018},
  pages = {2556--2565},
  doi = {10.1609/aaai.v32i1.11791}
}

@incollection{huber1992robust,
  title = {Robust estimation of a location parameter},
  author = {Huber, Peter J},
  booktitle = {Breakthroughs in statistics: Methodology and distribution},
  pages = {492--518},
  year = {1992},
  publisher = {Springer}
}

@inproceedings{8796030,
  author = {Ames, Aaron D. and Coogan, Samuel and Egerstedt, Magnus and Notomista, Gennaro and Sreenath, Koushil and Tabuada, Paulo},
  booktitle = {Eur. Control Conf. (ECC)},
  title = {Control Barrier Functions: Theory and Applications},
  year = {2019},
  pages = {3420-3431},
  doi = {10.23919/ECC.2019.8796030}
}

@article{ganaie2022ensemble,
  title = {Ensemble deep learning: A review},
  author = {Ganaie, Mudasir A and Hu, Minghui and Malik, Ashwani Kumar and Tanveer, Muhammad and Suganthan, Ponnuthurai N},
  journal = {Eng. Appl. Artif. Intell.},
  volume = {115},
  pages = {105151},
  year = {2022},
  publisher = {Elsevier}
}

@inproceedings{rame2021dice,
  title = {DICE: Diversity in Deep Ensembles via Conditional Redundancy Adversarial Estimation},
  author = {Alexandre Rame and Matthieu Cord},
  booktitle = {Proc. Int. Conf. Learn. Represent.},
  year = {2021}
}

@article{xu2025high,
  title = {High-Order Control Barrier Functions: Insights and a Truncated Taylor-Based Formulation},
  author = {Xu, Jianye and Alrifaee, Bassam},
  journal = {arXiv preprint arXiv:2503.15014},
  year = {2025}
}

@inproceedings{7524935,
  author = {Nguyen, Quan and Sreenath, Koushil},
  booktitle = {2016 American Control Conference (ACC)},
  title = {Exponential Control Barrier Functions for enforcing high relative-degree safety-critical constraints},
  year = {2016},
  pages = {322-328},
  doi = {10.1109/ACC.2016.7524935}
}

@article{chourasiya2026input,
  title = {An Input-Output Linearization-Based Robust Controller for Single-Track Models With Steering Actuator Delays},
  author = {Chourasiya, Sumit and Bascetta, Luca and Farina, Marcello and Ferretti, Gianni},
  journal = {IEEE Trans. Control Syst. Technol.},
  year = {2026},
  publisher = {IEEE}
}

@article{sale2023second,
  title = {Second-order uncertainty quantification: Variance-based measures},
  author = {Sale, Yusuf and Hofman, Paul and Wimmer, Lisa and H{\"u}llermeier, Eyke and Nagler, Thomas},
  journal = {arXiv preprint arXiv:2401.00276},
  year = {2023}
}

@misc{highwayenv,
  author = {Leurent, Edouard},
  title = {An Environment for Autonomous Driving Decision-Making},
  year = {2018},
  publisher = {GitHub},
  journal = {GitHub repository},
  howpublished = {\url{https://github.com/eleurent/highway-env}}
}

@article{idm,
  title = {Congested traffic states in empirical observations and microscopic simulations},
  journal = {Phys. Rev. E},
  author = {Treiber, Martin and Hennecke, Ansgar and Helbing, Dirk},
  year = {2002},
  month = {Jul},
  pages = {1805-1824},
  doi = {10.1103/physreve.62.1805}
}

@article{leng2025risk,
  title = {Risk-Aware Reinforcement Learning for Autonomous Driving: Improving Safety When Driving through Intersection},
  author = {Leng, Bo and Yu, Ran and Han, Wei and Xiong, Lu and Li, Zhuoren and Huang, Hailong},
  journal = {arXiv preprint arXiv:2503.19690},
  year = {2025}
}

@article{thananjeyan2021recovery,
  title = {Recovery rl: Safe reinforcement learning with learned recovery zones},
  author = {Thananjeyan, Brijen and Balakrishna, Ashwin and Nair, Suraj and Luo, Michael and Srinivasan, Krishnan and Hwang, Minho and Gonzalez, Joseph E and Ibarz, Julian and Finn, Chelsea and Goldberg, Ken},
  journal = {IEEE Robot. Autom. Lett.},
  volume = {6},
  number = {3},
  pages = {4915--4922},
  year = {2021},
  publisher = {IEEE}
}

@article{ma2021feasible,
  title = {Feasible actor-critic: Constrained reinforcement learning for ensuring statewise safety},
  author = {Ma, Haitong and Guan, Yang and Li, Shegnbo Eben and Zhang, Xiangteng and Zheng, Sifa and Chen, Jianyu},
  journal = {arXiv preprint arXiv:2105.10682},
  year = {2021}
}

@inproceedings{11423819,
  author = {Yu, Ran and Li, Zhuoren and Xiong, Lu and Han, Wei and Leng, Bo},
  booktitle = {2025 IEEE 28th Int. Conf. Intell. Transp. Syst. (ITSC)},
  title = {Uncertainty-Aware Safety-Critical Decision and Control for Autonomous Vehicles at Unsignalized Intersections},
  year = {2025},
  pages = {3805-3811},
  doi = {10.1109/ITSC60802.2025.11423819}
}

\end{document}